\documentclass[10pt,twocolumn,letterpaper]{article}
\PassOptionsToPackage{table}{xcolor}
\usepackage[pagenumbers]{cvpr}
\usepackage{amsthm,amsmath,graphicx,booktabs,tabularx,multirow,colortbl}
\newtheorem{proposition}{Proposition}
\definecolor{cvprblue}{rgb}{0.21,0.49,0.74}
\definecolor{oursrow}{HTML}{FFF0E6}
\definecolor{fullrow}{HTML}{F7F7F7}
\usepackage[pagebackref,breaklinks,colorlinks,allcolors=cvprblue]{hyperref}
\hypersetup{pdftitle={Gaussian-Reference Flow Guidance for Object Detection Dataset Distillation},
  pdfauthor={Kaixin Deng, Guang Li, Takahiro Ogawa, Miki Haseyama}}
\title{Gaussian-Reference Flow Guidance for Object Detection Dataset Distillation}
\author{
  Kaixin Deng\quad
  Guang Li\thanks{Correspondence to Guang Li
    \textless{}guang@lmd.ist.hokudai.ac.jp\textgreater{}.}\quad
  Takahiro Ogawa\quad
  Miki Haseyama\\[4pt]
  {\normalsize Hokkaido University}
}
\date{}
\begin{document}
\maketitle
\begin{abstract}
Dataset distillation for object detection compresses a large training set into a few annotated synthetic images. Recomposition methods reuse real instances with their exact annotations and achieve strong performance without pixel optimization. However, they limit the distilled appearance to the small fraction of instances selected under the budget. Introducing the appearance of the remaining instances through a generative model raises a second problem, which we call reference sharing. Reliable appearance statistics require large clusters, so many instances share each reference. A shared point target such as a cluster prototype or mode then homogenizes their appearance. To address these issues, we propose a two-stage framework that freezes the geometry of real instance blocks and re-renders their appearance with a frozen mask-conditioned flow model under Gaussian-Reference Flow Guidance (GFG). GFG estimates a diagonal Gaussian from each appearance cluster and applies it within the region of every aligned instance. In the guidance target, the cluster variance determines the weights of the cluster mean and the current state of the instance coordinate by coordinate. Instances that share a reference therefore receive different targets, point-reference guidance is the zero-variance limit, and the guidance stays bounded at the clean endpoint without a stop time. Experiments on COCO and PASCAL VOC show that our method achieves the highest AP at all evaluated budgets and transfers across four detector families.
\end{abstract}

\section{Introduction}
\label{sec:intro}

\begin{figure}[t]
  \centering
  \includegraphics[width=\linewidth]{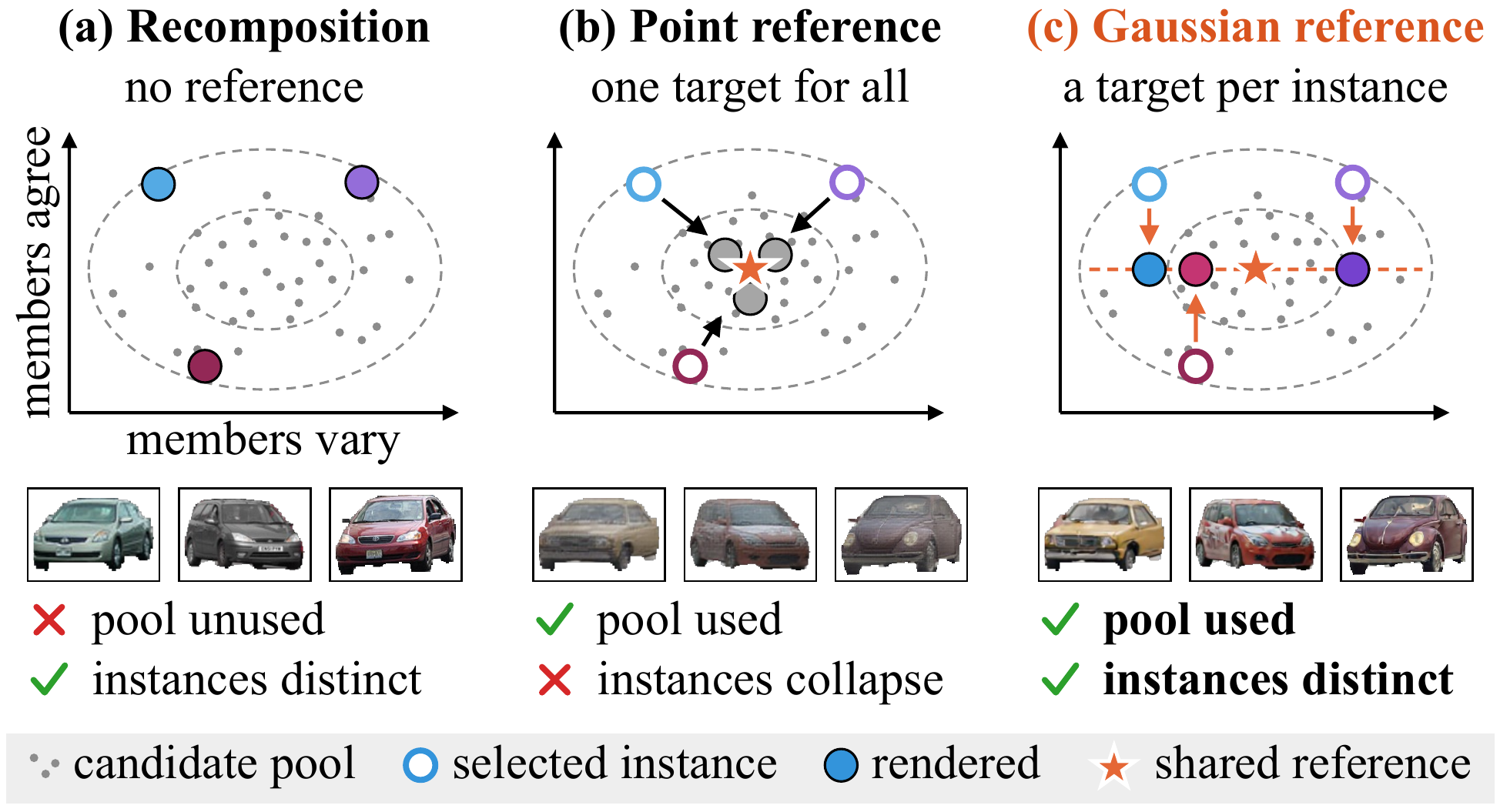}
  \caption{Concept of the proposed method. (a)~Recomposition leaves the candidate pool unused. (b)~A point reference collapses the instances that share it. (c)~A Gaussian reference draws each instance toward the cluster mean only where its members agree.}
  \label{fig:teaser}
\end{figure}

Dataset distillation~\cite{wang2018dataset,yu2023dataset} compresses a large training set into a small synthetic set on which models can be trained at a much lower cost. Most existing methods address image classification~\cite{zhao2021dataset,zhao2023dataset,cazenavette2022dataset,yin2023squeeze,sun2024diversity}, where each synthetic image carries a single label. Feature-matching methods can represent variation within a class through hyperbolic geometry~\cite{li2025hyperbolic} or statistics conditioned on sample difficulty~\cite{ma2026difficulty}. Object detection adds constraints: an image contains multiple instances, and the appearance and location of every instance must remain consistent with its category and bounding box. Existing detection distillation methods follow two routes. Synthesis-based methods invert a detector trained on the full dataset~\cite{qi2024fetch} or guide a diffusion model with task-specific proxies~\cite{qi2025towards}, and must keep the synthesized content consistent with every instance annotation. Recomposition methods such as OD$^3$~\cite{alkhatib2026od3} place real instances on new canvases, keep their exact annotations, and report strong performance without pixel optimization.

Although recomposition is effective, it limits the appearance of the distilled set to the selected instances (\Cref{fig:teaser}a). A small budget admits only a small fraction of the annotated instances, so the colors, textures and materials of all others are absent from the distilled images. As the budget decreases, each selected instance stands in for a larger part of the appearance distribution. This implies that the unselected instances carry appearance information that recomposition alone cannot use. 

Introducing this information through a generative model raises a second problem, which we call reference sharing. Generative distillation methods for classification summarize the data by points, such as the prototypes that initialize generation in D$^4$M~\cite{su2024d4m} and the modes that guide sampling in MGD$^3$~\cite{chansantiago2025mgd3}. In detection, the distilled set contains far more instances than there are appearance clusters with enough members to estimate statistics, so each reference serves many instances. A shared point target then pulls all assigned instances toward the same appearance and homogenizes them (\Cref{fig:teaser}b). 

Our method rests on a two-stage framework based on the principle \emph{freeze the geometry, re-render the appearance} (\Cref{fig:framework}). Geometry Distillation recomposes blocks of mutually touching real instances as wholes, preserving their relative layout and occlusion and fixing their annotations. Appearance Distillation then re-renders each block from its noised source pixels with a frozen generator~\cite{li2025seg2any} conditioned on the full region and category of every member. Re-rendering thus changes appearance without changing labels, but on its own it draws only on the source pixels and the generator prior, leaving the unselected instances unused.

To address both problems, we propose Gaussian-Reference Flow Guidance (GFG), a training-free guidance scheme. GFG estimates a diagonal Gaussian reference from each intra-class appearance cluster of the full candidate pool, aligns it to every instance of the cluster, and guides the instance toward the clean-latent prediction under this reference. This prediction combines the cluster mean and the current state coordinate by coordinate, with weights determined by the cluster variance. Unlike a point reference, it depends on the state of each instance, so instances that share a reference receive different targets (\Cref{fig:teaser}c), and point-reference guidance is its zero-variance limit. Its prediction tends to the current state at the clean endpoint, so the guidance velocity stays bounded and GFG runs at every step without a stop time. Appearance information from the unselected instances thus enters the distilled set without enlarging it or altering the fixed annotations. 
Our main contributions are summarized as follows:

\begin{itemize}
  \item We identify two obstacles to using the full instance population in detection distillation: recomposition limits the distilled appearance to the selected instances, and, under reference sharing, guidance toward a shared point target homogenizes the appearance of the many instances assigned to it, despite their differing source appearances.

  \item We propose a two-stage framework in which Geometry Distillation recomposes mutually touching instances as whole blocks with fixed layout, occlusion and annotations, and Appearance Distillation re-renders these blocks under GFG. To address reference sharing, GFG estimates a diagonal Gaussian reference for each appearance cluster of the full candidate pool, aligns it to every instance and fuses the resulting guidance region by region within each block at every sampling step.

  \item Experiments on COCO~\cite{lin2014microsoft} and PASCAL VOC~\cite{everingham2010pascal} show that our method achieves the highest AP at all evaluated budgets, and that its gains hold across four detector families with different underlying model architectures.
\end{itemize}

\section{Related Work}
\label{sec:related}

\noindent\textbf{Dataset Distillation for Object Detection.}
DCOD~\cite{qi2024fetch} transfers detection knowledge into model parameters and synthesizes images by model inversion, and UniDD~\cite{qi2025towards} guides diffusion with task-specific proxies across classification, detection and segmentation.
OD$^{3}$~\cite{alkhatib2026od3} avoids iterative pixel optimization through location-aware instance placement, observer-based screening and context extension.
Related work on classification builds synthetic images from foreground-aware crops~\cite{li2026foreground} or from discriminative patches of the same class~\cite{fan2026deco}. Fine-grained distillation also uses class prototypes to identify discriminative regions~\cite{ma2026fd2}. In detection, each retained region must remain consistent with its own box and with the surrounding instances.
Our method builds on this real-structure reuse, recomposing blocks of touching instances and re-rendering their appearance under fixed annotations.

\noindent\textbf{Representative Information in Generative Distillation.}
Generative distillation injects dataset information through the latent space of a generator~\cite{cazenavette2023generalizing}, initialization, model adaptation or sampling guidance.
Class information can enter the generator through text prompts~\cite{su2024diffusion}, paired visual and textual prototypes~\cite{zou2025vlcp}, or fusion of image and text features before synthesis~\cite{cai2026evlf}. Other approaches balance global structure and local detail during generator training~\cite{li2024global} or refine the colors of selected real images with a diffusion model~\cite{wu2025color}.
D$^{4}$M~\cite{su2024d4m} initializes diffusion from clustered latent prototypes, and Minimax Diffusion~\cite{gu2024efficient} adds training criteria for representativeness and diversity.
MGD$^{3}$~\cite{chansantiago2025mgd3} guides sampling toward discovered modes and then stops the guidance, IGD~\cite{chen2025influence} uses trajectory influence, and D$^{3}$HR~\cite{zhao2025taming} performs moment-matched group sampling after DDIM inversion~\cite{song2021denoising}.
Representation-guided diffusion uses prototypes from a self-supervised encoder to guide denoising~\cite{li2026srg}. Other classification methods track distribution alignment with an adaptive memory~\cite{li2025memory}, or select generated examples by classification difficulty~\cite{li2025sampling} or semantic relevance and diversity~\cite{li2026sas}. Detector-guided refinement checks the consistency of generated images with their class labels~\cite{zou2026label}. These methods target image classification, where each synthetic image can be associated with a single category.
In detection, however, the trade-off between reliable statistics and the number of instances forces many instances to share each reference.

\noindent\textbf{Structure-conditioned Generation and Flow Guidance.}
Instance composition~\cite{dwibedi2017cut,ghiasi2021simple,zhao2023xpaste,fan2024divergen} and diverse object sources improve detection and segmentation, and layout- and mask-conditioned generators~\cite{li2023gligen,zhang2023adding,chen2024geodiffusion,wang2024detdiffusion,zhu2024odgen} incorporate spatial annotations into synthesis.
Seg2Any~\cite{li2025seg2any} provides mask-, contour- and text-conditioned multi-instance generation on FLUX~\cite{flux2024,flux2025kontext}, which we use as a frozen backbone.
At sampling time, classifier and classifier-free guidance~\cite{dhariwal2021diffusion,ho2022classifier} guide generation toward a condition, and ILVR~\cite{choi2021ilvr} imposes low-frequency reference constraints.
Closest to our method, Follow the Mean~\cite{curvo2026follow} guides a frozen flow model toward the closed-form endpoint mean of a reference bank, a state-dependent kernel average of reference examples.
This average collapses onto the nearest example near the clean endpoint, and the correction is damped with a time schedule.
GFG instead estimates a diagonal Gaussian from a population cluster, so its prediction follows the current state on the coordinates where the cluster varies and its velocity stays bounded without a schedule or stop time.

% =====================================================================
\section{Method}
\label{sec:method}

\begin{figure*}[htb]
  \centering
  \includegraphics[width=\textwidth]{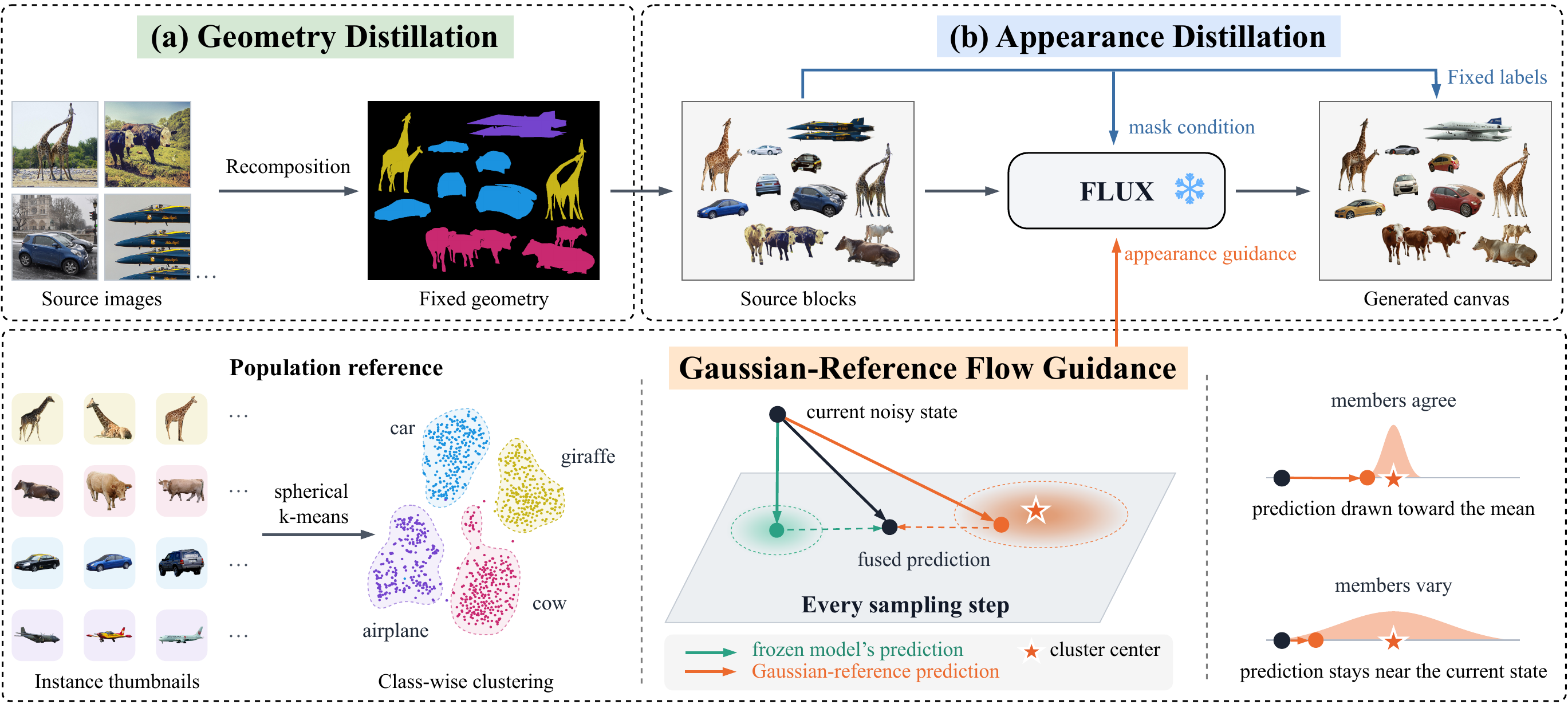}
  \caption{Overview of the proposed method. (a)~Geometry Distillation recomposes touching real instances onto canvases, which fixes the labels. (b)~Appearance Distillation re-renders each block with a frozen generator. At every sampling step, GFG fuses the Gaussian reference prediction of each class-wise cluster with the prediction of the frozen model.}
  \label{fig:framework}
\end{figure*}

Our framework separates detection dataset distillation into Geometry Distillation and Appearance Distillation (\Cref{fig:framework}).
In this section, we first formulate the task, then construct the fixed geometry and render it with a frozen generator, and finally introduce GFG.
The references of GFG are computed once from the full candidate pool, independently of selection and layout, so unselected candidates contribute without consuming budget.

% =====================================================================
\subsection{Problem Formulation}
\label{sec:formulation}

\paragraph{Task.}
Let $\mathcal{D}$ denote a detection training set of $N$ images with category, box and mask annotations for every instance, where missing masks are derived from the boxes.
We construct a synthetic set $\mathcal{S}$ of $M\ll N$ images, termed canvases, such that a detector trained on $\mathcal{S}$ approaches the performance of one trained on $\mathcal{D}$.
The index $i$ identifies an annotated instance with label $y_i$ in the category set $\mathcal{Y}$. 

\paragraph{Geometry and Rendering.}
For canvas $j$, the geometry $\mathcal{G}_j$ records each placed instance's category $y_i$, full region $F_i$ and visible region $V_i\subseteq F_i$.
Here, $F_i$ is the source mask after placement and before occlusion by other blocks, and $V_i$ is the part of $F_i$ not covered by instances placed in front of it.
Given the source pixels $I^{\mathrm{src}}_j$ of the placed instances, a renderer $\mathcal{R}$ and a label function $h$ produce
\begin{equation}
  \tilde I_j=\mathcal{R}\big(\mathcal{G}_j,I^{\mathrm{src}}_j\big),\qquad
  \tilde Y_j=h(\mathcal{G}_j),
  \label{eq:render}
\end{equation}
where $\tilde I_j$ and $\tilde Y_j$ denote the canvas image and its annotations, respectively, and $h$ assigns each instance its category $y_i$, visible mask $V_i$ and the tight bounding box of $V_i$.
Because $h$ depends only on $\mathcal{G}_j$, the renderer cannot change the annotations, which is the meaning of \emph{freeze the geometry, re-render the appearance}.
We call a renderer valid when the content rendered inside each $V_i$ keeps the category $y_i$ and fills the region, and our experiments examine this condition with two proxy metrics.
Three renderers share $\mathcal{G}_j$ and $I^{\mathrm{src}}_j$, so their differences are due to appearance alone: $\mathcal{R}_{\mathrm{copy}}$ pastes the source pixels, $\mathcal{R}_{\mathrm{regen}}$ re-renders them with a frozen generator, and $\mathcal{R}_{\mathrm{GFG}}$ adds population references.

\paragraph{Rectified Flow.}
We use the frozen Seg2Any generator~\cite{li2025seg2any}, a rectified flow model~\cite{liu2023flow,esser2024scaling} in the latent space of a variational autoencoder (VAE)~\cite{kingma2014auto,rombach2022high} with encoder $\mathrm{Enc}$.
Its condition $\kappa$ comprises per-instance region masks and texts, a contour image and a global prompt (details in the supplementary material).
A clean latent $x$ is noised along the linear path~\cite{liu2023flow,lipman2023flow}
\begin{equation}
  z_\sigma=(1-\sigma)\,x+\sigma\,\epsilon,\qquad \sigma\in[0,1],
  \label{eq:path}
\end{equation}
where $z_\sigma$ denotes the latent at noise level $\sigma$, and $\epsilon\sim\mathcal{N}(0,\mathbf{I})$ is Gaussian noise independent of $x$.
The network $v_\theta(z,\sigma;\kappa)$ with frozen parameters $\theta$ approximates the conditional mean of the path velocity $\epsilon-x$ given the current state.
Since $\epsilon-x=(z_\sigma-x)/\sigma$, it also gives the clean-latent prediction
\begin{equation}
  \hat x_\theta(z,\sigma)=z-\sigma\,v_\theta(z,\sigma;\kappa),
  \label{eq:net}
\end{equation}
which is defined for $\sigma>0$.

% =====================================================================
\subsection{Geometry Distillation}
\label{sec:geometry}

Geometry Distillation determines which source structures from a candidate pool, filtered by mask quality, scale and detector evidence, are retained and where they are placed.

\paragraph{Instance Blocks.}
Within each source image, candidates whose masks overlap or touch are connected, and each connected component, possibly a single instance, forms an instance block $b$.
Blocks stay indivisible during selection, placement and generation, so a member whose mask is cut by a touching neighbor keeps that neighbor, along with the observed layout and scale ratios.

\paragraph{Selection and Layout.}
Besides the canvas budget $M$, an instance budget $B$ limits the number of distinct source instances.
We greedily select complete blocks that most reduce a facility-location coverage error~\cite{nemhauser1978analysis,mirzasoleiman2020coresets} per source instance consumed, subject to per-category quotas.
Each selected block is uniformly scaled and translated as a whole, so its transformed masks define the full regions $F_i$.
The layout maximizes canvas coverage under visibility and size constraints, and occlusion between blocks follows their depth order.
The result is the fixed geometry $\mathcal{G}_j$ of every canvas (details in the supplementary material).

% =====================================================================
\subsection{Appearance Distillation}
\label{sec:appearance}

\paragraph{Source Initialization and Conditioning.}
Each block is rendered in a local window around its full regions.
The source anchor $I^{\mathrm{src}}_b$ contains the block's own pixels before cross-block occlusion, composited onto a neutral background.
Following the partial noising of SDEdit~\cite{meng2022sdedit}, we encode the anchor as $\mathrm{Enc}(I^{\mathrm{src}}_b)$, noise it along \cref{eq:path} to an intermediate level $\sigma_0\in(0,1)$, and start sampling there.
The anchor keeps the pose, viewpoint and appearance of the real instance, such as which side of an object is visible, which a silhouette alone does not determine, so the generator prior does not replace the variation of the data (see the supplementary material).
The noise level $\sigma_0$ trades this information against appearance change.
The condition $\kappa_b$ contains the full-region mask $F_i$ and category name of each member, the union of the member contours and a fixed global prompt, all expressed in the block window.
Conditioning on $F_i$ rather than $V_i$ describes the whole object, including occluded parts, so the visible fragment need not determine its appearance alone. In this way, occlusion introduced by recomposition does not truncate the object shape supplied to the generator.

\paragraph{Joint Rendering and Compositing.}
All members of a block are denoised jointly along one latent trajectory, and the decoded result is written to the canvas only within the union of their visible regions, so the annotations remain $h(\mathcal{G}_j)$.
This defines $\mathcal{R}_{\mathrm{regen}}$, whose appearance comes from the anchor and the prior of the frozen generator.
The generator is not adapted to the target dataset.
Therefore, $\mathcal{R}_{\mathrm{regen}}$ cannot aggregate appearance from unselected candidates, and GFG provides this information within the sampler.

% =====================================================================
\subsection{Gaussian-Reference Flow Guidance}
\label{sec:gfg}

GFG summarizes the candidate pool by Gaussian references, predicts from each the clean latent given the current state, and fuses the predictions with the network prediction.

\paragraph{Population References.}
Each candidate is cropped to its mask, placed on a neutral background and resized into a fixed canonical frame with its aspect ratio preserved.
The VAE encodes it into a canonical latent $u_i$ that is comparable across instances.
Within each category, spherical $k$-means~\cite{dhillon2001concept} clusters the normalized canonical latents into a number of clusters growing with the category size, and $C_i$ denotes the cluster of instance $i$.
Each cluster $C$ defines a Gaussian reference with mean $\mu_C$ and covariance $\operatorname{diag}(q_C)$, where $\mu_C$ and $q_C$ are the element-wise mean and variance of the unnormalized latents $u_i$ of its members, and each variance is floored at $q_{\min}>0$.
The variance thus describes the appearance variation among the members of a cluster over latent channels and spatial positions.

\paragraph{Gaussian Reference Prediction.}
Since references are shared, GFG conditions a shared reference on the state of each instance.
Omitting the cluster index, let $x\sim\mathcal{N}(\mu,\operatorname{diag}(q))$ denote the clean latent under the reference.
As in Follow the Mean~\cite{curvo2026follow}, the guidance target is the conditional mean of the clean latent given the current state, which Gaussian conditioning gives in closed form as follows:
\begin{equation}
  \bar x_G(z,\sigma)=(1-\omega)\odot\mu+\omega\odot\frac{z}{1-\sigma},
  \label{eq:ref_pred}
\end{equation}
where $\odot$ denotes element-wise multiplication, all other operations are also element-wise, and the weight
\begin{equation}
  \omega=\frac{(1-\sigma)^2q}{(1-\sigma)^2q+\sigma^2}
  \label{eq:omega}
\end{equation}
balances the reference mean against the current state for $\sigma\in(0,1)$.
By the same identity as in \cref{eq:net}, the reference velocity $v_G=(z-\bar x_G)/\sigma$ has the closed form
\begin{equation}
  v_G(z,\sigma)=\frac{\sigma(z-\mu)-(1-\sigma)\,q\odot z}{(1-\sigma)^2q+\sigma^2}.
  \label{eq:ref_vel}
\end{equation}

\begin{figure}[t]
  \centering
  \includegraphics[width=\linewidth]{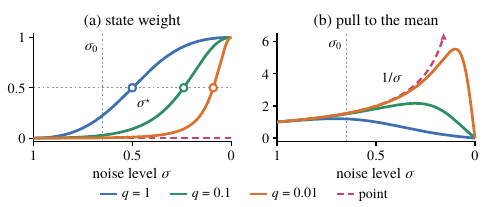}
  \caption{Behavior of the Gaussian reference flow. (a)~The weight $\omega$ of the current state rises as the noise level decreases, and each coordinate passes $\omega=1/2$ at its own release level $\sigma^\star$. (b)~The pull toward the mean, the weight of $z-\mu$ in $v_G$, stays bounded for $q>0$ and diverges as $1/\sigma$ for a point reference.}
  \label{fig:properties}
\end{figure}

\paragraph{Properties.}
\Cref{eq:ref_pred,eq:omega,eq:ref_vel} give GFG the following properties, which are stated and proved in the supplementary material.
First, the weight $\omega$ increases with the variance $q$, so coordinates on which the members of a cluster agree are drawn toward the mean, and coordinates on which they differ follow the current state.
Instances that share a reference therefore receive different targets instead of a shared point target, and this state dependence strengthens as the noise decreases. 
Second, as the variance vanishes, the reference prediction reduces to the cluster mean, so guidance toward point references such as cluster means and modes is the zero-variance limit of GFG.
Each coordinate is released from the mean at its own noise level $\sigma^\star=\sqrt{q}/(1+\sqrt{q})$, where $\omega=1/2$ (\Cref{fig:properties}a). Finally, as the noise level approaches zero, the reference velocity tends to $-z$ and remains bounded.
In contrast, the velocity of a point reference diverges (\Cref{fig:properties}b), and Follow the Mean~\cite{curvo2026follow} damps its correction with a time schedule.
Thus, GFG is applied at every step without a stop time. 

\paragraph{Instance Alignment and Fusion.}
For each member $i$ of a block, the affine map from the canonical frame to the block window resamples $(\mu_{C_i},q_{C_i})$ into aligned statistics $(\tilde\mu_i,\tilde q_i)$ (details in the supplementary material).
Applying \cref{eq:ref_pred,eq:omega,eq:ref_vel} to them at the corresponding coordinates of the block state gives $\bar x_{G,i}$ and $v_{G,i}$.
At each latent position, the weight $m_i\in[0,1]$ is derived from the fraction of the cell covered by $F_i$ and shared across channels.
Each weight vanishes outside the canonical support of its instance, and the weights of all block members are normalized to sum to at most one.
The fused clean-latent prediction is
\begin{equation}
  \hat x=\Big(1-\lambda\sum_{i\in b}m_i\Big)\odot\hat x_\theta+\lambda\sum_{i\in b}m_i\odot\bar x_{G,i},
  \label{eq:fusion}
\end{equation}
where $\lambda\in[0,1]$ controls the guidance strength, and all members share the block latent $z$ and the network prediction $\hat x_\theta$.
Each Euler step uses the velocity $(z-\hat x)/\sigma$, which equals
\begin{equation}
  v=v_\theta+\lambda\sum_{i\in b}m_i\odot\big(v_{G,i}-v_\theta\big).
  \label{eq:fusion_vel}
\end{equation}
When $\lambda=0$, GFG reduces to $\mathcal{R}_{\mathrm{regen}}$.
When $\bar x_{G,i}$ is replaced by $\tilde\mu_i$, it reduces to point-reference guidance.
GFG needs neither back-propagation nor additional training, and each step adds only element-wise operations per instance.

% =====================================================================
\section{Experiments}
\label{sec:exp}

% ---------------------------------------------------------------------
\subsection{Experimental Settings}
\label{sec:setup}

\begin{table*}[t]
\centering
\caption{Quantitative performance comparison on COCO and PASCAL VOC. Our method uses the same Faster R-CNN student architecture
and evaluation protocol as OD$^{3}$~\cite{alkhatib2026od3}.
Baseline results are taken from~\cite{qi2024fetch,qi2025towards,alkhatib2026od3}, and dashes denote results not reported there.
Full denotes the teacher trained on the full training set. Best results are in bold.}
\label{tab:main}
\footnotesize
\setlength{\tabcolsep}{3.0pt}
\renewcommand{\arraystretch}{1.08}
\begin{tabularx}{\textwidth}{@{}ll*{6}{>{\centering\arraybackslash}X}@{}}
\toprule
\multirow{2}{*}{\textbf{Budget}} &
\multirow{2}{*}{\textbf{Method}} &
\multicolumn{3}{c}{\textbf{COCO}} &
\multicolumn{3}{c}{\textbf{PASCAL VOC}} \\
\cmidrule(lr){3-5}\cmidrule(l){6-8}
& & \textbf{AP} & \textbf{AP$_{50}$} & \textbf{AP$_{75}$}
& \textbf{mAP} & \textbf{AP$_{50}$} & \textbf{AP$_{75}$} \\
\midrule

% ==================== 0.5% ====================
\multirow{8}{*}{$0.5\%$}
& Random
& $5.5 \pm 0.1$ & $14.2 \pm 0.1$ & $2.9 \pm 0.1$
& $4.9 \pm 0.2$ & $15.8 \pm 0.7$ & $1.4 \pm 0.1$ \\

& Uniform
& $5.6 \pm 0.1$ & $14.3 \pm 0.1$ & $2.9 \pm 0.1$
& $5.0 \pm 0.2$ & $15.8 \pm 0.5$ & $1.5 \pm 0.1$ \\

& K-Center
& $2.8 \pm 0.0$ & $8.9 \pm 0.1$ & $0.7 \pm 0.1$
& $3.8 \pm 0.3$ & $14.5 \pm 0.6$ & $0.9 \pm 0.1$ \\

& Herding
& $2.6 \pm 0.1$ & $8.8 \pm 0.1$ & $0.8 \pm 0.1$
& $3.6 \pm 0.0$ & $12.6 \pm 0.2$ & $0.9 \pm 0.0$ \\

& DCOD~\cite{qi2024fetch}
& $10.0 \pm 0.1$ & $21.5 \pm 0.2$ & $8.0 \pm 0.1$
& $14.2 \pm 0.5$ & $37.9 \pm 0.9$ & $6.6 \pm 0.4$ \\

& UniDD~\cite{qi2025towards}
& $7.1 \pm 0.4$ & $16.9 \pm 0.3$ & --
& $8.5 \pm 0.4$ & $22.3 \pm 0.6$ & -- \\

& OD$^{3}$~\cite{alkhatib2026od3}
& $17.2 \pm 0.1$ & $31.9 \pm 0.2$ & $16.9 \pm 0.1$
& -- & $38.5 \pm 0.1$ & -- \\

\rowcolor{oursrow}
& \textbf{GFG (ours)}
& $\boldsymbol{19.9 \pm 0.3}$ & $\boldsymbol{34.9 \pm 0.3}$ & $\boldsymbol{20.5 \pm 0.0}$
& $\boldsymbol{20.0 \pm 0.3}$ & $\boldsymbol{47.2 \pm 0.6}$ & $\boldsymbol{13.2 \pm 0.2}$ \\

\midrule

% ==================== 1% ====================
\multirow{8}{*}{$1\%$}
& Random
& $8.3 \pm 0.0$ & $19.7 \pm 0.1$ & $5.3 \pm 0.1$
& $8.6 \pm 0.5$ & $25.5 \pm 0.7$ & $3.0 \pm 0.2$ \\

& Uniform
& $8.4 \pm 0.1$ & $19.7 \pm 0.2$ & $5.4 \pm 0.1$
& $8.8 \pm 0.4$ & $25.7 \pm 0.6$ & $3.2 \pm 0.3$ \\

& K-Center
& $4.0 \pm 0.1$ & $12.9 \pm 0.1$ & $1.2 \pm 0.1$
& $6.1 \pm 0.2$ & $21.9 \pm 0.9$ & $1.4 \pm 0.1$ \\

& Herding
& $4.1 \pm 0.1$ & $12.5 \pm 0.2$ & $1.3 \pm 0.2$
& $5.5 \pm 0.2$ & $19.3 \pm 0.5$ & $1.3 \pm 0.1$ \\

& DCOD~\cite{qi2024fetch}
& $12.1 \pm 0.1$ & $24.7 \pm 0.2$ & $10.4 \pm 0.1$
& $19.8 \pm 0.4$ & $46.4 \pm 0.4$ & $13.0 \pm 0.6$ \\

& UniDD~\cite{qi2025towards}
& $10.8 \pm 0.4$ & $22.5 \pm 0.5$ & --
& $16.8 \pm 0.5$ & $38.9 \pm 0.7$ & -- \\

& OD$^{3}$~\cite{alkhatib2026od3}
& $22.4 \pm 0.1$ & $\boldsymbol{39.5 \pm 0.2}$ & $22.9 \pm 0.1$
& -- & $51.1 \pm 0.2$ & -- \\

\rowcolor{oursrow}
& \textbf{GFG (ours)}
& $\boldsymbol{23.9 \pm 0.2}$ & $39.0 \pm 0.2$ & $\boldsymbol{26.0 \pm 0.1}$
& $\boldsymbol{24.6 \pm 0.5}$ & $\boldsymbol{52.8 \pm 0.7}$ & $\boldsymbol{19.4 \pm 0.9}$ \\

% ==================== Full Teacher ====================
\midrule
\rowcolor{fullrow}
Full &
\shortstack[l]{Faster R-CNN}
& $39.8$ & $60.1$ & $43.3$
& $59.3$ & $85.8$ & $66.1$ \\

\bottomrule
\end{tabularx}
\end{table*}

\paragraph{Datasets and Budgets.}
We distill COCO train2017 and the PASCAL VOC 2007 and 2012 trainval sets, and evaluate on COCO val2017 and the VOC 2007 test set. As VOC provides masks for only part of its images, we obtain every object mask with SAM~\cite{kirillov2023segment} prompted by its ground-truth box. The image budgets are $0.5\%$ and $1\%$ of the training images, with the canvas numbers $M$ and instance budgets $B$ in the supplementary material. Candidate selection, clustering and reference estimation use training data only. 

\paragraph{Metrics.}
Following OD$^3$~\cite{alkhatib2026od3}, we report box AP, AP$_{50}$ and AP$_{75}$ on COCO and mAP, AP$_{50}$ and AP$_{75}$ on VOC, where AP and mAP average over IoU thresholds from $0.5$ to $0.95$ and VOC uses area-based precision--recall integration.
We further characterize the distilled instances with four instance-level metrics.
Diversity and coverage use the cosine distance between unit-normalized features of an independent encoder.
For every evaluated cluster $C$, let $\tilde{\mathcal{H}}_C$ be its $n_C\ge2$ generated instances and $\mathcal{H}_C$ a fixed set of its real instances that are not among the source instances.
Diversity $\mathrm{Div}_C$ is the mean pairwise distance within $\tilde{\mathcal{H}}_C$, and coverage $\mathrm{CovDist}_C$ is the mean distance from each real instance in $\mathcal{H}_C$ to its nearest generated instance (lower is better), both macro-averaged over the same clusters for all strategies.
Because Div also increases when rendering fails, two further metrics serve as validity checks.
Category recognizability (Valid) is the fraction of instances whose category an independent vision-language model recognizes inside the annotated region, and mIoU measures the agreement between independently estimated object masks and the fixed visible masks, with both protocols in the supplementary material. These metrics serve as proxies for rendering validity as defined in \Cref{sec:formulation}, rather than predictors of downstream training utility.

\paragraph{Students and Protocol.}
We follow the evaluation protocol of OD$^{3}$~\cite{alkhatib2026od3} without modification.
The student detector is trained on the distilled set with the same schedule and with feature distillation from the same fixed teacher trained on the full training set (details in the supplementary material). Baseline results in \Cref{tab:main} are taken from their respective source papers~\cite{qi2024fetch,qi2025towards,alkhatib2026od3} and retain their original evaluation protocols.
For the OD$^{3}$ results in \Cref{tab:cross,tab:components,fig:vis,fig:classwise}, which require additional students or per-category and per-image predictions, we train the students on the released OD$^{3}$ data~\cite{alkhatib2026od3} under the same protocol. Faster R-CNN~\cite{ren2015faster} is the default student, and we further evaluate cross-detector transfer with RetinaNet~\cite{lin2017focal}, RT-DETRv2~\cite{zhao2024detrs,lv2024rtdetrv2} and YOLO26~\cite{jocher2026yolo26} under the same protocol. \Cref{tab:main} reports results over multiple seeds, and the other tables report a single run with a fixed seed shared by all rows.

\paragraph{Implementation Details.}
We use the official Seg2Any LoRA~\cite{hu2022lora} for FLUX.1-dev~\cite{flux2024} with all weights frozen, sample every block with 32 Euler steps from $\sigma_0=0.65$, and set $\lambda=0.25$ and $q_{\min}=10^{-4}$.

\paragraph{Comparative Methods.}
We compare with coreset selection (Random, Uniform, K-Center~\cite{sener2018active} and Herding~\cite{welling2009herding}) and with the detection distillation methods DCOD~\cite{qi2024fetch}, UniDD~\cite{qi2025towards} and OD$^3$~\cite{alkhatib2026od3}.
Unless noted, internal comparisons fix our geometry and change only the renderer of \cref{eq:render}: \emph{Copy} ($\mathcal{R}_{\mathrm{copy}}$), \emph{Regen} ($\mathcal{R}_{\mathrm{regen}}$) and \emph{GFG} ($\mathcal{R}_{\mathrm{GFG}}$).

% ---------------------------------------------------------------------
\subsection{Main Results}
\label{sec:main}

\paragraph{Benchmark Comparison.}
\Cref{tab:main} shows that our method achieves the highest AP among the reported results on both datasets at both budgets.
At the $1\%$ budget, it reaches $60\%$ of the full-data AP on COCO and $41\%$ on VOC.
On COCO, it exceeds the strongest prior method, OD$^3$, by $2.7$ and $1.5$ AP at the two budgets, well beyond the standard deviation over seeds.
The gain is larger at the strict IoU threshold, where AP$_{75}$ improves by more than $3$ points at both budgets, whereas AP$_{50}$ is slightly below that of OD$^3$ at the $1\%$ budget.
On VOC, our method leads on every reported metric.
On both datasets, the gain over OD$^3$ shrinks as the budget grows, probably because each selected instance then stands in for fewer unselected instances.

\begin{table}[t]
\centering
\small
\setlength{\tabcolsep}{2.5pt}
\renewcommand{\arraystretch}{1.12}
\caption{Box AP of cross-detector transfer on COCO at the $0.5\%$ budget. The same distilled set and protocol are used for every student.}
\label{tab:cross}
\begin{tabularx}{\columnwidth}{@{}l*{4}{>{\centering\arraybackslash}X}@{}}
\toprule
Method
& \begin{tabular}{@{}c@{}}Faster\\R-CNN\end{tabular}
& RetinaNet
& \begin{tabular}{@{}c@{}}RT-\\DETRv2\end{tabular}
& YOLO26 \\
\midrule
OD$^3$ & $17.4$ & $18.7$ & $18.1$ & $13.9$ \\
\rowcolor{oursrow}
\textbf{GFG (ours)} & $\textbf{20.1}$ & \textbf{21.0} & \textbf{21.2} & $\textbf{17.1}$ \\
\bottomrule
\end{tabularx}
\par\smallskip
\end{table}

\paragraph{Cross-architecture Generalization.}
\Cref{tab:cross} reports four detector families trained on the same distilled sets.
GFG improves over OD$^3$ with every student, by $2.3$ to $3.2$ AP, so its benefit is not specific to the default student.

\begin{figure*}[t]
  \centering
  \includegraphics[width=\textwidth]{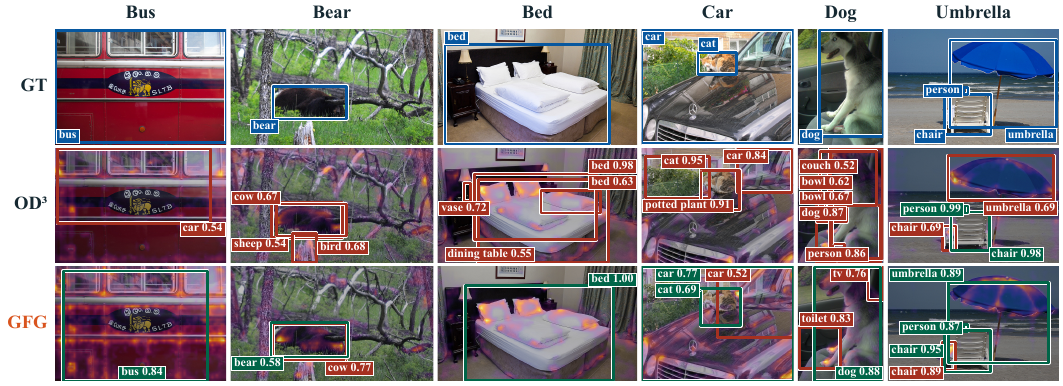}
  \caption{Qualitative comparison on COCO val2017. The rows show the ground truth and the detections with Grad-CAM maps of the Faster R-CNN students trained on the OD$^3$ and GFG sets, where correct and incorrect detections are shown in green and red.}
  \label{fig:vis}
\end{figure*}

\paragraph{Qualitative Comparison.}
To examine how the additional appearance affects the students, \Cref{fig:vis} compares the Faster R-CNN students of OD$^3$ and GFG on six randomly sampled validation images.
The OD$^3$ student confuses categories, repeats boxes and covers only part of some objects that it labels correctly.
In most examples, the Grad-CAM~\cite{selvaraju2017grad} maps of both students fall on the objects, which suggests that GFG improves how the attended regions are recognized and delimited, in line with its broader appearance coverage.

% ---------------------------------------------------------------------
\subsection{Ablation Study}
\label{sec:gains}

\begin{table}[t]
\centering
\small
\setlength{\tabcolsep}{4pt}
\caption{Ablation of the construction unit and the renderer on COCO at the $0.5\%$ budget with Faster R-CNN. The last three rows share the same geometry and labels and differ only in the renderer.}
\label{tab:components}
\begin{tabular}{@{}llccc@{}}
\toprule
Construction unit & Renderer & AP & AP$_{50}$ & AP$_{75}$ \\
\midrule
OD$^3$~\cite{alkhatib2026od3} & -- & 17.4 & 31.0 & 16.8 \\
\midrule
Independent instances & Copy  & 13.1 & 23.9 & 13.0 \\
Real instance blocks  & Copy  & 18.7 & 34.2 & 18.4 \\
Real instance blocks  & Regen & 18.5 & 34.2 & 18.1 \\
\rowcolor{oursrow}
Real instance blocks  & \textbf{GFG}   & \textbf{20.1} & \textbf{35.2} & \textbf{20.6} \\
\bottomrule
\end{tabular}
\end{table}

\paragraph{From Recomposition to Population-Guided Rendering.}
\Cref{tab:components} adds one component at a time.
Independent instances and real instance blocks share the source instances, category counts, canvases, background and instance budget, with matched scales and visible-area distributions.
Under Copy, blocks improve over independent instances by $5.6$ AP, the largest single increment, which isolates the value of keeping the observed structure of touching instances.
Copy on blocks is already above OD$^3$, so the geometry alone is competitive with the strongest prior method.
On the same geometry and labels, Regen stays at the level of Copy, and GFG adds $1.6$ AP over Regen.
Re-rendering with the frozen generator alone therefore does not help the student, because it adds no dataset information beyond the source pixels. The gain comes from the population references, which contribute the appearance statistics of the full candidate pool.
The two stages also act at different IoU thresholds.
Copy already provides most of the AP$_{50}$ gain over OD$^3$, whereas most of the AP$_{75}$ gain appears only with GFG.
Thus, the block geometry mainly improves detection at the loose threshold, and the references mainly tighten localization, probably because the varied appearance inside the fixed masks helps the detector fit object extents. 

\begin{table}[t]
  \centering
  \small
  \setlength{\tabcolsep}{4pt}
  \caption{Ablation of the source of the reference statistics on COCO at the $0.5\%$ budget with Faster R-CNN. The cluster partition, geometry, anchors and seeds are fixed. CovDist is multiplied by 100.}
  \label{tab:ablation}
  \begin{tabular}{@{}lccc@{}}
    \toprule
    Statistics & AP$\uparrow$ & CovDist$\downarrow$ & mIoU$\uparrow$ \\
    \midrule
    Selected instances only   & 19.3 & 18.6 & \textbf{87.2} \\
    Unselected instances only & 20.0 & 18.1 & 87.0 \\
    \rowcolor{oursrow}
    Full candidate pool (\textbf{GFG}) & \textbf{20.1} & \textbf{18.0} & 87.1 \\
    \bottomrule
  \end{tabular}
\end{table}

\paragraph{Effect of the Reference Statistics Source.}
\Cref{tab:ablation} changes only the instances from which the reference statistics are estimated.
Statistics of the selected instances alone recover about half of the gain of the full pool over Regen, whereas those of the unselected instances alone recover nearly all of it.
CovDist decreases in the same order. Thus, the unselected instances, which recomposition leaves unused, provide most of the gain.

% ---------------------------------------------------------------------
\subsection{Analysis}
\label{sec:gaussian}

\begin{table*}[t]
  \centering
  \small
  \caption{Comparison of strategies under shared references on COCO at the $0.5\%$ budget with Faster R-CNN. All rows use the same layout. Div and CovDist are multiplied by 100. Valid and mIoU are validity checks, and Copy gives their reference level on real pixels.}
  \label{tab:strategies}
  \begin{tabular}{llccccc}
    \toprule
    Strategy & Initialization & AP$\uparrow$ & Div & CovDist$\downarrow$ & Valid$\uparrow$ & mIoU$\uparrow$ \\
    \midrule
    Real instance blocks (Copy) & --  & 18.7  & 26.4 & 18.9  & \textbf{67.3}  &  87.3 \\ 
    \midrule
    Seg2Any, unguided           & Pure noise       & 15.2 & 28.1 & 20.8 & 59.5 & 87.8 \\
    Regen                       & Real-block RGB   & 18.5  & 26.4  & 19.1  & 67.0 &  87.0 \\ 
    D$^4$M & Cluster prototype & 5.9 & 20.9 & 25.1 & 33.2 & 85.6 \\
    MGD$^3$ & Real-block RGB   & 12.4 & 22.8 & 21.2 & 47.9 & \textbf{89.1} \\
    \rowcolor{oursrow}
    \textbf{GFG}         & Real-block RGB   & \textbf{20.1} & 26.0 & \textbf{18.0} & 65.8 & 87.1 \\
    \bottomrule
  \end{tabular}
  \par\smallskip
\end{table*}

\begin{figure}[t]
  \centering
  \includegraphics[width=\linewidth]{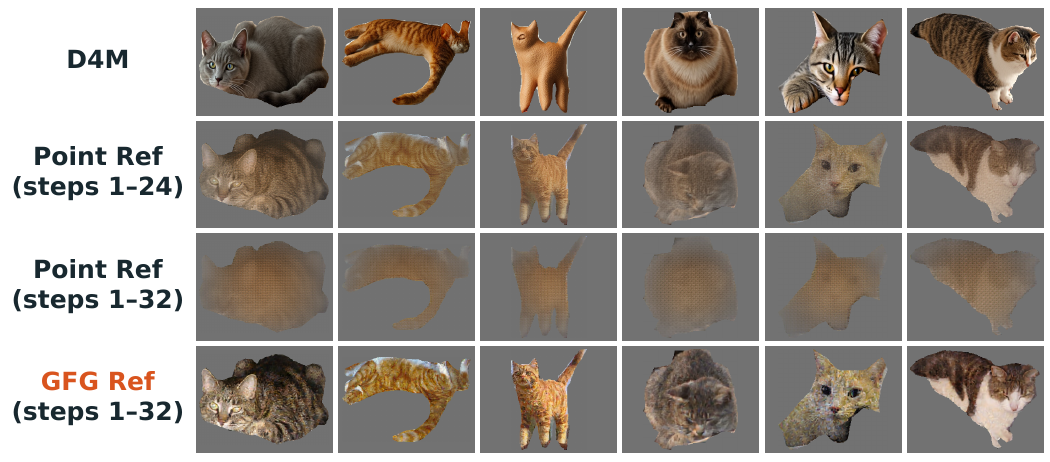}
  \caption{Instances that share a population reference. The rows show D$^4$M-style initialization, MGD$^3$-style point-reference guidance over steps 1 to 24 and 1 to 32, and GFG. The 32-step point reference renders the cats alike, whereas GFG keeps them distinct.}
  \label{fig:shared}
\end{figure}

\paragraph{Strategies.}
To test the effect of reference sharing, \Cref{tab:strategies} compares strategies that share a reference among many instances with Copy, Regen and unguided generation.
The D$^4$M-style and MGD$^3$-style baselines are mechanism-level controls implemented in our detection framework, and they do not reproduce the original classification methods~\cite{su2024d4m,chansantiago2025mgd3}. Further comparisons are provided in the supplementary material.
D$^4$M-style initialization replaces the anchor by the spatially mapped cluster prototype plus noise.
MGD$^3$-style guidance keeps the anchor and guides every instance toward the aligned cluster mean with the same strength $\lambda=0.25$ as GFG, so its full variant over all 32 steps is the zero-variance limit of GFG.
\Cref{tab:strategies} reports the variant that stops after step 24, and \Cref{fig:shared} also shows the full one.
All generative strategies share layouts, conditions, cluster assignments, noise, steps and student training, so they differ only in the initialization and the guidance.

\begin{figure}[t]
  \centering
  \includegraphics[width=0.92\linewidth]{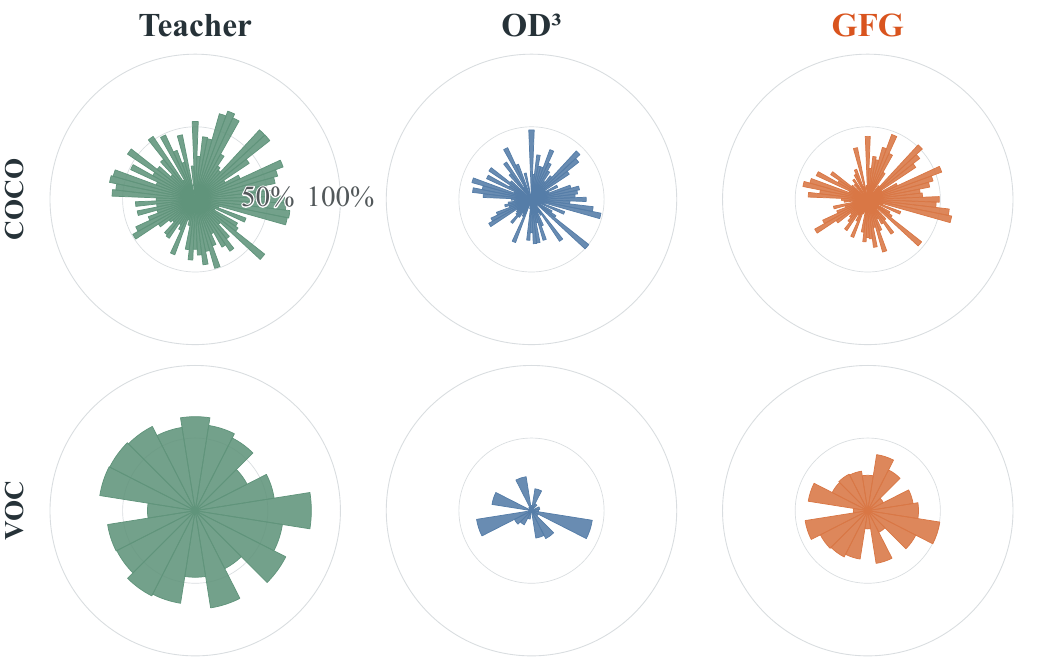}
  \caption{Per-category AP of the teacher, OD$^3$ and GFG on COCO and VOC. Each wedge denotes one category.}
  \label{fig:classwise}
\end{figure}

\paragraph{Shared References.}
\Cref{tab:strategies} shows that the D$^4$M-style and MGD$^3$-style baselines both lower Div below that of the real instances and AP far below that of Copy.
In \Cref{fig:shared}, point-reference guidance over all 32 steps renders the cats alike, whereas GFG, which differs from it only in the reference variance, keeps them distinct.
The homogenization is therefore due to the point target.
Stopping at step 24 reduces the attraction and recovers part of their structure, but it does not make the target instance-specific.
The stopped baseline remains 7.7 AP below GFG.
GFG stays close to the real instances in Div, Valid and mIoU, suggesting that its generated instances remain recognizable and consistent with the fixed masks. Notably, even Copy, which uses real pixels with correct labels, achieves only 67.3 Valid. We therefore treat Valid as an auxiliary reference rather than a definitive measure of rendering validity.
Starting from the source anchor instead of pure noise raises both Valid and AP, which supports the role of the anchor in providing pose and viewpoint.
Across the eight settings of \Cref{tab:ablation,tab:strategies}, the ranking by AP follows the ranking by CovDist, which measures how far the unselected real instances lie from the generated ones.
Div does not follow it, as pure noise gives the highest Div but a lower AP than Copy.
Neither do Valid and mIoU once rendering is valid: Copy and Regen have higher Valid than GFG but lower AP, and the MGD$^3$-style baseline has the highest mIoU but a much lower AP, whereas Valid does separate the failed renderers (D$^4$M-style and MGD$^3$-style).
Together, these findings indicate that shared references should capture within-cluster variation.

\paragraph{Category-wise Trend.}
To examine whether the students follow the category trend of the full-data teacher, \Cref{fig:classwise} shows the per-category AP of all three.
Both students broadly follow the teacher's trend, but GFG follows it more closely: on VOC, the Spearman correlation between the per-category AP of the student and that of the teacher rises from $0.14$ for OD$^3$ to $0.55$ for GFG.

\section{Conclusion}
In this work, we identify that recomposition leaves the unselected instances unused and that, under reference sharing, a point target homogenizes the generated instances.
To address these issues, we propose a two-stage framework that freezes the geometry of real instance blocks and re-renders their appearance under Gaussian-Reference Flow Guidance (GFG).
GFG estimates a diagonal Gaussian reference from each appearance cluster of the full candidate pool, and the cluster variance gives different targets to the instances that share a reference.
Experiments on COCO and PASCAL VOC show that our method achieves the highest AP at all evaluated budgets, and its gains on COCO hold across four detector families.
Limitations are discussed in Appendix~\ref{sec:supp_limitations}.
Future work will extend the reference model beyond diagonal Gaussians and improve the rendering of small objects.

{\small
\bibliographystyle{ieeenat_fullname}
\bibliography{main}
}

\clearpage
\appendix
\section*{Appendix}
\setcounter{equation}{0}
\setcounter{table}{0}
\setcounter{figure}{0}
\renewcommand{\theequation}{S\arabic{equation}}
\renewcommand{\thetable}{S\arabic{table}}
\renewcommand{\thefigure}{S\arabic{figure}}
\renewcommand{\theproposition}{S\arabic{proposition}}
\providecommand{\theHequation}{}\renewcommand{\theHequation}{S\arabic{equation}}
\providecommand{\theHtable}{}\renewcommand{\theHtable}{S\arabic{table}}
\providecommand{\theHfigure}{}\renewcommand{\theHfigure}{S\arabic{figure}}

This supplementary material provides the details and analyses referred to in the main paper.
We first detail Geometry Distillation and Appearance Distillation.
We then derive the properties of GFG and describe its implementation.
Finally, we provide the experimental details, additional results and the limitations of our method.
Equations, tables and figures numbered with the prefix S belong to this supplementary material, and all other numbers refer to the main paper.

% =====================================================================
\section{Geometry Distillation Details}
\label{sec:supp_geometry}

This section details how Geometry Distillation builds the candidate pool, forms instance blocks, selects them under the instance budget and places them on the canvases.

\subsection{Candidate Pool}
\label{sec:supp_candidates}

We select the candidate pool per category from the valid non-crowd instances of the training set according to the detector evidence, scale and mask quality.
The detector evidence is a class-agnostic check with the Faster R-CNN teacher of the same dataset.
We remove an instance only when no teacher detection of any category matches it under loose thresholds on confidence and box IoU.
This check discards the instances that the teacher does not detect as any object and keeps those whose category it confuses.

The remaining instances receive three scores.
The resolution score of instance $i$ is defined as
\begin{equation}
  \gamma_i=\frac{1}{2}\min\Big(1,\frac{l_i}{32}\Big)+\frac{1}{2}\min\Big(1,\sqrt{\frac{|\Omega_i|}{512}}\Big),
  \label{eq:supp_resolution}
\end{equation}
where $\Omega_i$ denotes the pixel set of its source mask, and $l_i$ denotes the short side of the tight box of this mask.
The score equals one when the instance reaches the minimum size required on the canvases, namely 32 pixels on the short side and 512 pixels in area, without upsampling.
The structure score is the fraction of the mask that lies in its largest 8-connected component.
The border score is one if the mask does not touch the image border and zero otherwise, because an instance cut by the border would appear truncated on a canvas.
We rank the instances of each category by the mean of the three scores.
Let $\bar n_y$ denote the number of remaining instances of category $y$.
The candidate pool keeps the $n_y$ top-ranked instances of this category, where
\begin{equation}
  n_y=\min\big(\bar n_y,\max(1000,\lceil \bar n_y/2\rceil)\big).
  \label{eq:supp_pool}
\end{equation}
A category with at most 1000 remaining instances thus keeps all of them, and a category with more than 2000 keeps its top half.
In this way, the cut-off of each category is set by rank and not by a fixed score threshold.

\subsection{Instance Blocks}
\label{sec:supp_blocks}

Within each source image, we connect two candidates $i$ and $i'$ when their masks overlap or touch in the 8-neighborhood.
The edge set $E$ of this graph is defined as
\begin{equation}
  (i,i')\in E\iff\min_{w\in\Omega_i,\;w'\in\Omega_{i'}}\lVert w-w'\rVert_\infty\le 1,
  \label{eq:supp_block_edge}
\end{equation}
where $w$ and $w'$ denote pixel coordinates.
The instance blocks are the connected components of this graph.
Since the graph contains only candidates, an instance excluded from the pool can split a group of touching instances into several blocks.

\paragraph{Indivisibility of Blocks.}
We keep each block whole during selection, placement and generation for three reasons.
First, splitting a block would separate a truncated object from the cause of its truncation.
The masks of COCO and the SAM masks of VOC cover only the visible part of each object, and Geometry Distillation performs no amodal completion.
Therefore, the mask of an object occluded by a touching neighbor is cut where the neighbor covers it.
If such an instance were placed alone, the cut would appear without an occluder, and its box would enclose a truncated shape.
Within its block, the occluder is placed beside it, so the truncation remains consistent with the rendered scene.
Truncations caused by excluded or unannotated objects remain, because these occluders belong to no block.

Second, touching instances can interact, such as a person riding a horse or holding a racket.
A block keeps these members at their observed relative positions and scales, so the canvases retain the co-occurrence and relative scale of real scenes.
In contrast, independent placement scales and places every member separately.

Third, touching instances are common.
On COCO, the $426{,}445$ candidates form $234{,}410$ blocks, and $153{,}360$ of these blocks contain a single instance.
The remaining $273{,}085$ candidates, about $64\%$ of the pool, touch at least one other candidate and form $81{,}050$ blocks with $3.4$ members on average.
Treating instances independently would therefore separate most candidates from a neighbor.

Table~3 of the main paper isolates the effect of this choice.
With the same source instances, instance budget and renderer, blocks improve over independent instances by $5.6$ AP.
The cost is that a block consumes the budget of all its members, so the selection normalizes the coverage gain of a block by its number of members.

\subsection{Category Quotas}
\label{sec:supp_quota}

We divide the instance budget $B$ into category quotas that mix a frequency-proportional allocation with a uniform one.
The quota of category $y$ is first set to
\begin{equation}
  B_y=B\,\Big(0.8\,\frac{n_y}{\sum_{y'\in\mathcal{Y}}n_{y'}}+\frac{0.2}{|\mathcal{Y}|}\Big),
  \label{eq:supp_quota}
\end{equation}
where $n_y$ denotes the number of candidates of category $y$.
We then apply capacity-constrained redistribution and largest-remainder rounding, so the quotas are integers that sum to $B$.
The number of selected instances of category $y$ must stay within a tolerance $\delta_y=\max(5,\lceil0.1B_y\rceil)$ of its quota.

\subsection{Coverage Objective and Greedy Selection}
\label{sec:supp_selection}

We measure coverage with DINOv3 features, where $f_i$ denotes the feature of instance $i$.
For each category $y$, we draw a probe set $\mathcal{P}_y$ of $256$ candidates, stratified over class-wise clusters of these features.
Each probe $r$ carries a weight $\beta_r$, and the weights restore the cluster populations and sum to one within each category.
These feature clusters serve only for selection and are distinct from the appearance clusters of GFG.
The coverage error of a set $\mathcal{I}$ of selected instances is defined as
\begin{equation}
  \mathcal{L}_{\mathrm{cov}}(\mathcal{I})=\sum_{y\in\mathcal{Y}}\frac{B_y}{B}\sum_{r\in\mathcal{P}_y}\beta_r\,d_r(\mathcal{I}),
  \label{eq:supp_coverage}
\end{equation}
where $d_r(\mathcal{I})$ denotes the distance from probe $r$ to its nearest selected instance of the same category.
For a probe of category $y$, this distance is given by
\begin{equation}
  d_r(\mathcal{I})=\min\Big(2,\min_{i\in\mathcal{I},\,y_i=y}\big(1-\cos(f_r,f_i)\big)\Big),
  \label{eq:supp_dist}
\end{equation}
so it takes its maximum value of two when category $y$ has no selected instance.

Let $\Delta(b\mid\mathcal{I})=\mathcal{L}_{\mathrm{cov}}(\mathcal{I})-\mathcal{L}_{\mathrm{cov}}(\mathcal{I}\cup b)$ denote the coverage gain of block $b$.
Starting from the empty set, each greedy step selects the block
\begin{equation}
  b^\star=\arg\max_{b\in\mathcal{A}(\mathcal{I})}\;\frac{\Delta(b\mid\mathcal{I})}{|b|},
  \label{eq:supp_greedy}
\end{equation}
where $|b|$ denotes the number of members of block $b$, and $\mathcal{A}(\mathcal{I})$ denotes the set of admissible blocks.
A block is admissible when it is unselected, fits the remaining budget, reuses no source instance, respects the upper bounds of the quotas and keeps their lower bounds reachable.
Normalizing by $|b|$ prices a block by the source instances that it consumes, so a large block is selected only if it adds proportionally more coverage.
At each step, we retrieve the candidate blocks from the feature clusters with the largest remaining coverage error.
Ties are broken by the improvement in quota agreement, and a few exchanges of whole blocks repair the remaining quota deficits.

\paragraph{Submodularity.}
The coverage error is a facility-location objective, and we verify that its reduction $2-\mathcal{L}_{\mathrm{cov}}$ is monotone and submodular over blocks.
The quantity $\phi_r(\mathcal{I})=2-d_r(\mathcal{I})$ can be written as
\begin{equation}
  \phi_r(\mathcal{I})=\max\Big(0,\max_{i\in\mathcal{I},\,y_i=y}\big(1+\cos(f_r,f_i)\big)\Big),
  \label{eq:supp_phi}
\end{equation}
where the similarity $1+\cos(f_r,f_i)$ lies in $[0,2]$, and the maximum over an empty set is taken as $-\infty$.
Because the quota weights $B_y/B$ and the probe weights $\beta_r$ each sum to one, we obtain
\begin{equation}
  2-\mathcal{L}_{\mathrm{cov}}(\mathcal{I})=\sum_{y\in\mathcal{Y}}\frac{B_y}{B}\sum_{r\in\mathcal{P}_y}\beta_r\,\phi_r(\mathcal{I}).
  \label{eq:supp_facility}
\end{equation}
The right-hand side is a non-negative combination of facility-location functions with non-negative similarities.
Each term is monotone and submodular in $\mathcal{I}$, and so is their combination.
A block contributes the union of its instances, and composing a monotone submodular function with this union preserves both properties.
Therefore, the objective is also monotone and submodular over sets of blocks.

\subsection{Shared Transformation}
\label{sec:supp_scale}

We transform each block as a whole by a uniform scale $s$ and a translation, so its members keep their relative layout.
The scale is restricted to the interval $[s_{\min}(b),s_{\max}(b)]$.
The lower limit is given by
\begin{equation}
  s_{\min}(b)=\max_{i\in b}\max\Big(\frac{32}{l_i},\sqrt{\frac{512}{|\Omega_i|}}\Big),
  \label{eq:supp_smin}
\end{equation}
which keeps every member at least 32 pixels on its short side and 512 pixels in area.
The upper limit is given by
\begin{equation}
  s_{\max}(b)=\min\Big(2,\frac{1024}{W_b},\frac{1024}{H_b}\Big),
  \label{eq:supp_smax}
\end{equation}
where $W_b$ and $H_b$ denote the width and height of block $b$ in its source image, respectively.
This limit restricts upsampling to a factor of two and fits the block into a canvas of $1024\times1024$ pixels.
Before selection, we exclude the blocks with an empty interval and the blocks with a mask that covers at least $99.5\%$ of its source image.
When a selected block fails the exact rasterized check at its realized scale, we replace it as a whole by blocks chosen with the same rule.

\subsection{Visibility Constraints}
\label{sec:supp_visibility}

As in the main paper, $F_i$ denotes the full region of instance $i$, namely its source mask after the block transformation and before occlusion by other blocks.
Let $b(i)$ denote the block of instance $i$, and let $U_b=\bigcup_{i\in b}F_i$ denote the union of the full regions of block $b$.
The depth rank of block $b$ is denoted by $\pi_b$, where a larger rank is closer to the viewer.
The visible region of instance $i$ is given by
\begin{equation}
  V_i=F_i\setminus\bigcup_{b':\,\pi_{b'}>\pi_{b(i)}}U_{b'},
  \label{eq:supp_visible}
\end{equation}
and its occlusion ratio is $o_i=1-|V_i|/|F_i|$.
Since no amodal completion is performed, the occlusion within a block is the one observed in the source image.
A layout is admissible only if the occlusion ratio of every instance does not exceed a limit $\tau_i$.
In addition, every visible region must contain at least 512 pixels and have a tight box with a short side of at least 32 pixels.
It must also remain non-empty on the 16-pixel attention grid of the generator.
We set $\tau_i=0.25$ and relax it to $\tau_i=0.5$ for the instances whose full region covers at least $20\%$ of the canvas.

\subsection{Placement}
\label{sec:supp_placement}

We assign the blocks to the $M$ canvases by balancing the estimated mask area and the number of instances per canvas.
The blocks that remain large at their minimum scale are placed first, with at most one of them per canvas.
For a block at a given scale, we rank the candidate positions by the cross-correlation between its mask and the occupied canvas on a grid with a stride of 8 pixels.
We test a placement in front of and behind the placed blocks against the visibility constraints and accept the admissible position with the largest gain in the foreground union.
This rule realizes the maximization of canvas coverage described in the main paper.
After all blocks are placed, we enlarge the block scales in up to three passes whenever the foreground union grows and all constraints still hold.

\subsection{Size-aware Ordering}
\label{sec:supp_order}

With the positions, scales and full regions fixed, we re-optimize the depth order to protect small instances.
The objective is given by
\begin{equation}
  \min_{\pi}\;\sum_i \eta_i\,o_i(\pi),
  \label{eq:supp_order}
\end{equation}
where $o_i(\pi)$ denotes the occlusion ratio of instance $i$ under the depth order $\pi$, and the weight $\eta_i=\operatorname{clip}\big(96/\sqrt{|F_i|},1,8\big)$ increases as the instance becomes smaller.
The search is subject to the visibility constraints.
In addition, no instance with a full region of at most $32^2$ pixels may lose visible area.
The mean occlusion of the instances with a full region of at most $96^2$ pixels may not increase on any canvas.
The two areas follow the COCO size thresholds.
We use a bounded local search with insertions and swaps and accept only strict improvements.

\subsection{Statistics}
\label{sec:supp_geometry_stats}

\begin{figure*}[t]
  \centering
  \includegraphics[width=\linewidth]{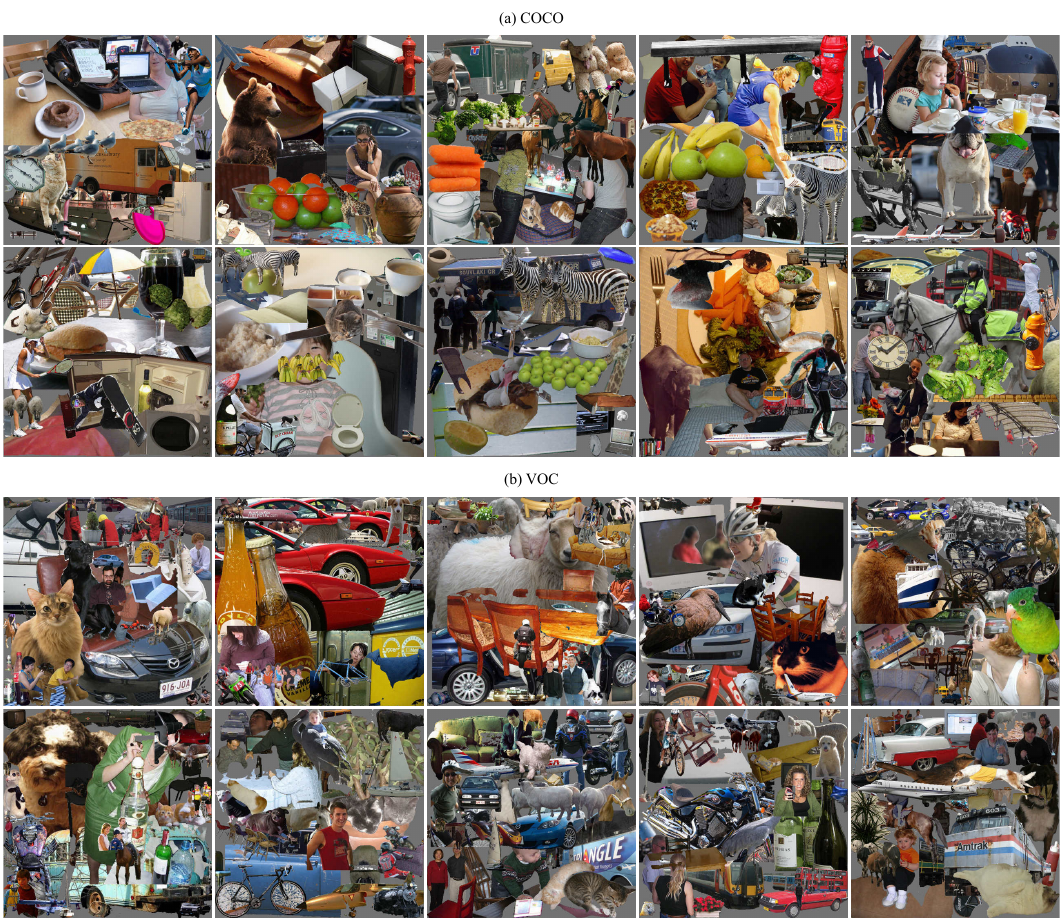}
  \caption{Visualization of the distilled canvases of GFG at the $0.5\%$ budget. (a)~Ten canvases on COCO. (b)~Ten canvases on VOC. Each canvas is composed of real instance blocks whose appearance is re-rendered under GFG, and the pixels outside all visible regions keep the gray background.}
  \label{fig:supp_images}
\end{figure*}

\begin{figure*}[t]
  \centering
  \includegraphics[width=\linewidth]{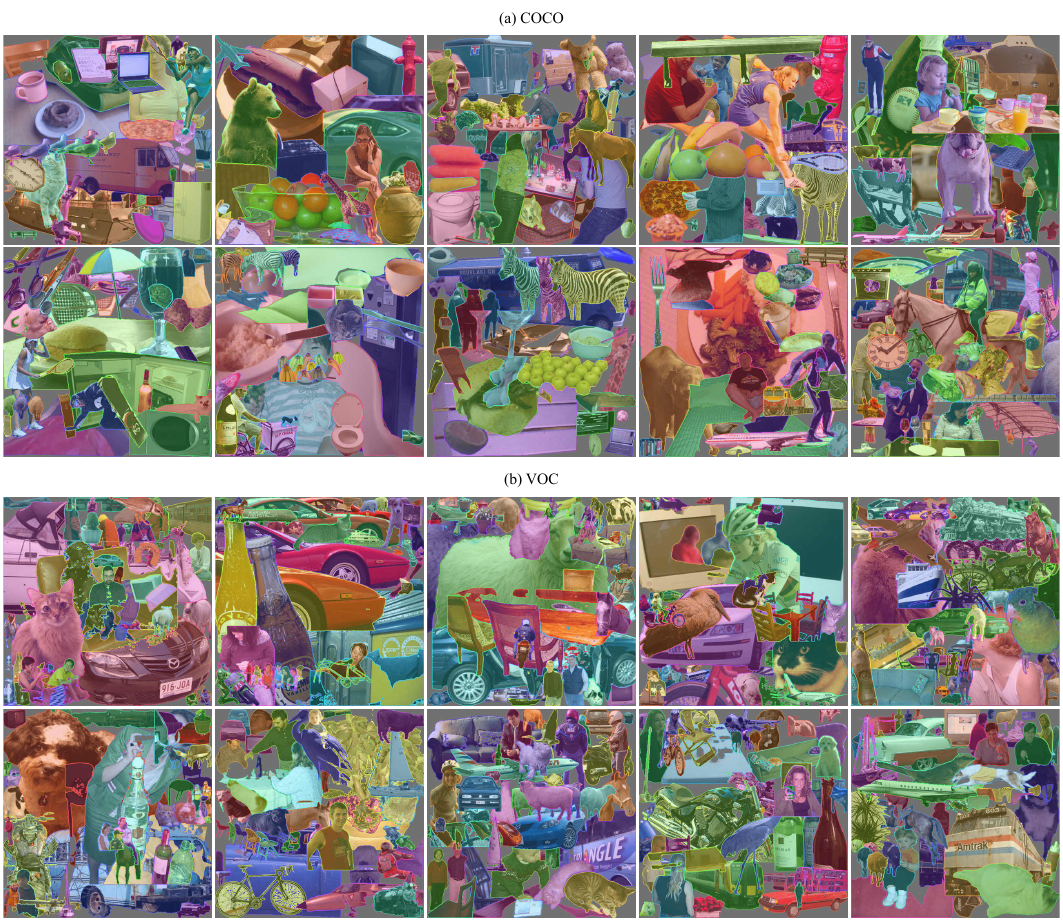}
  \caption{Fixed annotations of the canvases in \Cref{fig:supp_images}. Each color denotes the visible mask $V_i$ of one instance after the block transformation and the cross-block occlusion, and the tight box of $V_i$ is its detection label. (a)~The COCO masks are provided by the dataset. (b)~The VOC masks are obtained with SAM prompted by the ground-truth boxes, so only box annotations are required.}
  \label{fig:supp_masks}
\end{figure*}

On COCO train2017, the candidate pool keeps $426{,}445$ of the $849{,}941$ valid non-crowd instances.
These candidates form $234{,}410$ blocks, and $81{,}050$ of these blocks contain more than one instance.
At the $1\%$ budget, Geometry Distillation yields $17{,}493$ blocks with $50{,}000$ instances on $1{,}184$ canvases.
Every selected instance appears exactly once, and the mean foreground coverage of the canvases is $88.1\%$.
The rasterized check replaces $495$ blocks by $634$ blocks with the same number of instances, and the size-aware ordering changes the depth order on $1{,}036$ of the $1{,}184$ canvases.
\Cref{tab:supp_budgets} lists the numbers of canvases and the instance budgets of all settings.
\Cref{fig:supp_images} shows ten distilled canvases of each dataset, and \Cref{fig:supp_masks} shows their fixed annotations.

% The two figures below show the same canvases in the same order.

% =====================================================================
\section{Appearance Distillation Details}
\label{sec:supp_seg2any}

This section details the conditioning inputs of the generator, the block window, the source anchor and the write-back, and examines the effect of the source initialization.

\subsection{Conditioning Inputs}
\label{sec:supp_seg2any_inputs}

We use the official Seg2Any LoRA weights for FLUX.1-dev, which were trained on SACap-1M, and keep all weights frozen.
Seg2Any separates a segmentation condition into regional semantics and region shapes.
For the semantics, an attention mask binds the image tokens of each region to the tokens of its text.
In addition, the image tokens of different regions do not attend to one another in image self-attention.
For the shapes, the region boundaries enter as a contour image that the VAE encodes into additional condition tokens.
We use the hard attention mode, which applies these restrictions in transformer blocks 19 to 37 at every sampling step.
\Cref{tab:supp_seg2any} lists the inputs for a block $b$.
The first four inputs form the condition $\kappa_b$ of the main paper, and the last one is the initial latent of the sampler.

\begin{table}[t]
  \centering
  \small
  \caption{Seg2Any inputs and their settings for a block $b$.}
  \label{tab:supp_seg2any}
  \begin{tabularx}{\columnwidth}{@{}l>{\raggedright\arraybackslash}X@{}}
    \toprule
    Input & Setting \\
    \midrule
    Region masks   & Full regions $F_i$ of the members of $b$ on the window grid \\
    Region texts   & Category name of each member \\
    Global prompt  & Fixed sentence (\cref{sec:supp_synthesis}) \\
    Contour image  & Union of the four-neighbor inner contours of the members, at half the window resolution \\
    Initial latent & Noised anchor, \cref{eq:supp_anchor_init} \\
    \bottomrule
  \end{tabularx}
\end{table}

\subsection{Block Window, Source Anchor and Write-back}
\label{sec:supp_anchor}
\label{sec:supp_window}

The block window is the bounding box of the full regions of the block members, enlarged by 32 pixels, clipped to the canvas and padded to a multiple of 32 pixels.
Its side length lies between 256 and 1024 pixels.
We choose the offset of the padding such that every member remains non-empty on the 16-pixel attention grid of the generator.

We reconstruct the source anchor $I^{\mathrm{src}}_b$ from the source pixels of the block before cross-block occlusion.
The pixels are resampled by foreground-normalized bilinear interpolation within the union of the source masks, so the source background does not bleed into the instances.
All remaining pixels are set to a constant gray as the neutral background.
Since Seg2Any has no native argument for an input image or its strength, we pass the noised anchor as the initial latent.
This latent is given by
\begin{equation}
  z_{\sigma_0}=(1-\sigma_0)\,\mathrm{Enc}\big(I^{\mathrm{src}}_b\big)+\sigma_0\,\epsilon,
  \label{eq:supp_anchor_init}
\end{equation}
where $\epsilon\sim\mathcal{N}(0,\mathbf{I})$ denotes Gaussian noise, and $\mathrm{Enc}$ returns the posterior mode of the FLUX VAE after its latent shift and scaling.
This initialization is the partial noising along the linear path of Eq.~(2) of the main paper.
A larger $\sigma_0$ reduces the contribution of the anchor, and $\sigma_0=1$ corresponds to sampling from pure noise.

After decoding, we write only the pixels inside the union of the visible regions of the block back to the canvas.
The canvas pixels outside all visible regions keep the gray background.
This write-back keeps the annotations $h(\mathcal{G}_j)$ of Eq.~(1) of the main paper.
However, it cannot undo a change of viewpoint inside a region.
The three renderers of the main paper share the write-back.
Copy writes the source pixels of the anchor directly.
Regen samples from the initial latent of \cref{eq:supp_anchor_init} with $\lambda=0$, and GFG samples from the same initial latent under the guidance of Eq.~(8) of the main paper.

\subsection{Effect of Source Initialization}
\label{sec:supp_viewpoint}

\begin{figure*}[t]
  \centering
  \includegraphics[width=\linewidth]{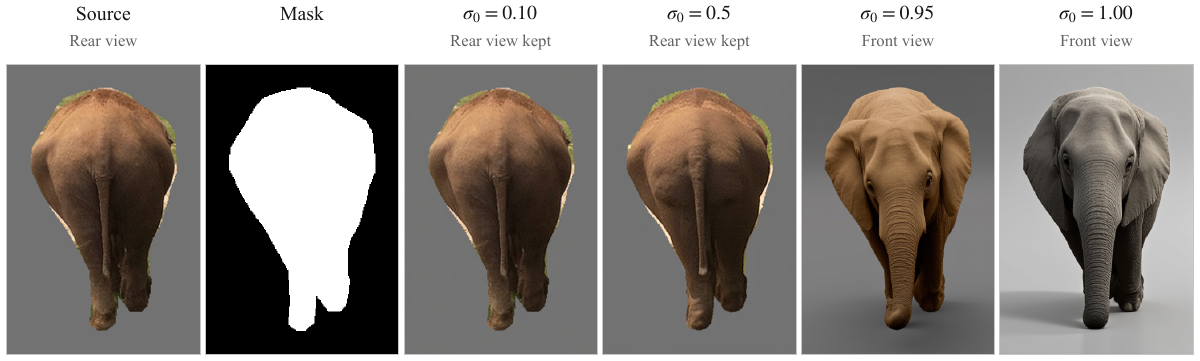}
  \caption{Effect of the initial noise level $\sigma_0$ on the re-rendered appearance. The source and its mask are followed by the outputs at four noise levels, where $\sigma_0=1.00$ starts from pure noise. The outputs share the noise, the conditions and 32 Euler steps and use no reference guidance. The low noise levels keep the rear view and the brownish coat of the source, whereas the high noise levels yield a frontal view inside the same silhouette. All panels share the same mask-derived crop and show the decoded outputs before the write-back.}
  \label{fig:supp_viewpoint}
\end{figure*}

To examine why sampling starts from the source anchor instead of pure noise, we compare different initial noise levels $\sigma_0$ on one instance.
As stated in the main paper, the anchor keeps the pose, viewpoint and appearance of the real instance, which a silhouette alone does not determine.
As shown in \Cref{fig:supp_viewpoint}, the source shows an elephant from behind with a brownish coat.
The outputs at $\sigma_0=0.10$ and $\sigma_0=0.5$ keep this rear view.
At $\sigma_0=0.95$, a frontal elephant with eyes, ears and a trunk appears inside the same silhouette.
At $\sigma_0=1.00$, the coat also turns to the gray of a typical elephant.
Thus, the generator prior replaces both the viewpoint and the color of the real instance when the anchor is weak, and the write-back keeps this change inside the labeled region.

The same effect appears at the dataset level in Table~5 of the main paper.
Sampling from pure noise gives the highest Div but also a higher CovDist and a lower AP than Copy and Regen.
This suggests that the added variation comes from the generator prior and not from the data.
With the anchor and without reference guidance, Regen stays at the level of Copy in AP, Div and CovDist, so the rendered instances remain close to their sources.
These results indicate that the anchor preserves the variation of the real instances, and the variation of the unselected instances enters only through GFG.

% The figure shows sigma_0 = 0.5, not the default 0.65, so its panel label must not say "default".
% If the panel is regenerated at 0.65, replace 0.5 by 0.65 in the text and in the caption.

\paragraph{Experimental Settings.}
The source is a COCO training instance of the category elephant.
We center it on a gray canvas of $1024\times1024$ pixels and resize its longest side to 768 pixels with the aspect ratio preserved.
The region text is \texttt{elephant}, and the global prompt is the fixed sentence used for all blocks.
The four outputs share the anchor latent, the Gaussian noise, the mask, the text and the contour tokens.
The seeds of the noise and the contour are 20261031 and 20261032, respectively.
Each trajectory uses 32 Euler steps from its initial noise level to zero, where the network runs in BF16 and the latent updates are computed in FP32.
We disable the reference guidance by setting $\lambda=0$ to isolate the effect of the initialization, and the mask, contour and text conditions remain active at every noise level.

% =====================================================================
\section{Gaussian-Reference Flow Guidance Details}
\label{sec:supp_gfg}

This section details the population references, proves the properties of the Gaussian reference prediction, and describes the instance alignment and the spatial weights of GFG.

\subsection{Canonical Latents and Clusters}
\label{sec:supp_clusters}

We crop each candidate to the tight box of its mask and composite it onto the gray background.
The crop is resized with its aspect ratio preserved to fit a square of $256\times256$ pixels with a margin of 16 pixels.
The FLUX VAE encodes this image into the canonical latent $u_i\in\mathbb{R}^{16\times32\times32}$.
Within each category, spherical $k$-means clusters the normalized vectors $\mathrm{vec}(u_i)/\lVert\mathrm{vec}(u_i)\rVert_2$.
The number of clusters of category $y$ is given by
\begin{equation}
  K_y=\min\Big(n_y,\;128,\;\max\big(4,\lceil n_y/500\rceil\big)\Big),
  \label{eq:supp_num_clusters}
\end{equation}
which targets about 500 members per cluster and caps the number of clusters per category at 128.
The statistics of a cluster $C$ are computed from the unnormalized latents of its members.
The mean is given by
\begin{equation}
  \mu_C=\frac{1}{|C|}\sum_{i\in C}u_i,
  \label{eq:supp_mean}
\end{equation}
and the variance is given by
\begin{equation}
  q_C=\max\Big(\frac{1}{|C|}\sum_{i\in C}(u_i-\mu_C)^2,\;q_{\min}\Big),
  \label{eq:supp_stats}
\end{equation}
where the square and the maximum are element-wise, and $q_{\min}$ denotes the variance floor.

As shown in \Cref{tab:supp_budgets}, each reference is shared by many instances.
On COCO at the $1\%$ budget, the $50{,}000$ selected instances fall into $784$ of the $787$ clusters, with about $64$ instances per cluster.
Moreover, $88\%$ of the candidates that define the references and $94\%$ of all valid non-crowd instances never appear on a canvas.
Across the four settings, a cluster serves 32 to 75 selected instances on average, and $76\%$ to $94\%$ of the candidates never appear on a canvas.
If the number of clusters matched the number of selected instances, the $426{,}445$ candidates would be split into clusters of fewer than $9$ members on average.
Such clusters are too small to estimate a mean and a variance in a latent space of $16{,}384$ dimensions.
For this reason, many instances have to share each reference, as discussed in the main paper.

\subsection{Derivation of the Reference Prediction}
\label{sec:supp_proof}

The properties of GFG stated in the main paper follow from standard Gaussian conditioning along the linear path.
\Cref{prop:supp_reference} states them formally in the order of the main paper.

\begin{proposition}[State dependence and limits]
\label{prop:supp_reference}
Let the clean latent follow $x\sim\mathcal{N}(\mu,\operatorname{diag}(q))$ with $q>0$, let $z_\sigma$ be defined by Eq.~(2) of the main paper, and let $\sigma\in(0,1)$.
Then $\bar x_G(z,\sigma)=\mathbb{E}[x\mid z_\sigma=z]$ is given by Eq.~(4) of the main paper with the weight $\omega$ of Eq.~(5), and the following properties hold.
(i)~For two states $z$ and $z'$, $\bar x_G(z,\sigma)-\bar x_G(z',\sigma)=\omega\odot(z-z')/(1-\sigma)$, where $\omega$ increases with $q$ and as $\sigma$ decreases.
(ii)~As $q\to0$, $\bar x_G\to\mu$ for every $z$.
(iii)~The weight satisfies $\omega=1/2$ exactly at $\sigma^\star=\sqrt{q}/(1+\sqrt{q})$, with $\omega<1/2$ for $\sigma>\sigma^\star$ and $\omega>1/2$ for $\sigma<\sigma^\star$.
(iv)~For a fixed $z$, $\bar x_G\to z$ and the reference velocity $v_G=(z-\bar x_G)/\sigma\to-z$ as $\sigma\to0$, whereas the point-reference velocity $(z-\mu)/\sigma$ diverges unless $z=\mu$.
\end{proposition}

\begin{proof}
Let $a=1-\sigma\in(0,1)$.
Each coordinate of $(x,z_\sigma)$ is jointly Gaussian with $\mathbb{E}[z_\sigma]=a\mu$, $\mathrm{Var}(z_\sigma)=a^2q+\sigma^2$ and $\mathrm{Cov}(x,z_\sigma)=aq$.
Gaussian conditioning therefore gives
\begin{equation}
  \mathbb{E}[x\mid z_\sigma=z]=\mu+g\odot(z-a\mu),
  \label{eq:supp_conditioning}
\end{equation}
where $g=aq/(a^2q+\sigma^2)$.
Since $ag=\omega$, the right-hand side equals $(1-\omega)\odot\mu+\omega\odot z/a$, which is Eq.~(4) of the main paper.

For (i), the difference follows because $\bar x_G$ is affine in $z$.
Writing $\omega=1/\big(1+\sigma^2/(a^2q)\big)$ shows that $\omega$ increases with $q$ and decreases with $\sigma/(1-\sigma)$, which increases with $\sigma$.
For (ii), we have $\omega\to0$ as $q\to0$ for a fixed $\sigma\in(0,1)$, and hence $\bar x_G\to\mu$.
For (iii), Eq.~(5) of the main paper gives $\omega=1/2$ if and only if $\sigma/(1-\sigma)=\sqrt{q}$, which is equivalent to $\sigma=\sigma^\star$, and the monotonicity in (i) gives the two inequalities.

For (iv), we have $\omega\to1$ and $a\to1$ as $\sigma\to0$ for a fixed $q>0$, and hence $\bar x_G\to z$.
Since $\epsilon-x=(z_\sigma-x)/\sigma$, the reference velocity is the conditional mean of the path velocity under the reference, namely $v_G=\mathbb{E}[\epsilon-x\mid z_\sigma=z]=(z-\bar x_G)/\sigma$.
We have $z-\bar x_G=(1-\omega)\odot(z-\mu)-\sigma\,\omega\odot z/a$.
Using $1-\omega=\sigma^2/(a^2q+\sigma^2)$ and $\omega/a=aq/(a^2q+\sigma^2)$, we obtain Eq.~(6) of the main paper.
As $\sigma\to0$, the numerator of Eq.~(6) tends to $-q\odot z$ and its denominator tends to $q$, so $v_G\to-z$.
In contrast, $(z-\mu)/\sigma$ diverges unless $z=\mu$.
\end{proof}

Property (iii) implies that a coordinate is drawn mainly toward the mean while the noise level is above $\sigma^\star$ and mainly follows the current state afterwards.
Coordinates with a small variance are released late, and coordinates with a large variance are released early, so every coordinate obtains its own release level from the spread of its cluster.

\subsection{Endpoint Behavior of the Correction}
\label{sec:supp_velocity}

GFG is implemented in the velocity form of Eq.~(8) of the main paper, which follows from the prediction form of Eq.~(7) with $\hat x=z-\sigma v$ and $\hat x_\theta=z-\sigma v_\theta$.
We refer to the term $\lambda\sum_{i\in b}m_i\odot(v_{G,i}-v_\theta)$ as the correction.
By \cref{prop:supp_reference}(iv), the correction remains bounded as the noise level approaches zero whenever $v_\theta$ is bounded.
In contrast, the correction of a point reference grows as $1/\sigma$.

The bound depends on the variance.
Since $(1-\sigma)^2q+\sigma^2\ge2\sigma(1-\sigma)\sqrt{q}$ holds by the inequality of arithmetic and geometric means, Eq.~(6) of the main paper gives
\begin{equation}
  |v_G|\le\frac{1}{1-\sigma}\Big(\frac{|z-\mu|}{2\sqrt{q}}+|z|\Big)
  \label{eq:supp_vel_bound}
\end{equation}
for every coordinate and every $\sigma\in(0,1)$.
The coefficient of $z-\mu$ in Eq.~(6) is the pull toward the mean shown in Figure~3b of the main paper, and it is largest near the release level of the coordinate.
At the floor $q_{\min}=10^{-4}$, this coefficient reaches about $50$ near $\sigma=0.01$.
At $\sigma=0.03$, it equals $30.2$, close to the value of $33.3$ for a point reference.
A floored coordinate is released only at $\sigma^\star\approx0.01$.
If this level lies below the smallest positive noise level of the sampler, the coordinate is guided much like a point reference throughout sampling.
Coordinates with a larger variance are released inside the sampling range, for example at $\sigma^\star\approx0.09$ for $q=0.01$ and at $\sigma^\star=0.5$ for $q=1$.
GFG departs from a point reference on these coordinates.

Whether the correction also vanishes at the clean endpoint depends on the data.
Let $\rho_\sigma$ denote the density of $z_\sigma$.
The ideal velocity of any data distribution is given by
\begin{equation}
  \mathbb{E}[\epsilon-x\mid z_\sigma=z]=-\frac{z+\sigma\nabla_z\log \rho_\sigma(z)}{1-\sigma},
  \label{eq:supp_ideal_vel}
\end{equation}
which follows from Tweedie's formula.
This velocity tends to $-z$, the limit of $v_G$, when $\sigma\nabla_z\log \rho_\sigma(z)$ vanishes.
In this case, the correction of an accurate $v_\theta$ fades at the clean endpoint.
For data concentrated near low-dimensional structures, this term need not vanish along the trajectory, and the correction then remains bounded without fading.
Applying GFG at every sampling step requires only the boundedness.

\subsection{Instance Alignment and Spatial Weights}
\label{sec:supp_alignment}

Let $\psi_i$ denote the affine map from the canonical frame of instance $i$ to its block window.
It composes the inverse of the canonicalization, the placement of the instance in the layout and the crop of the window.
Resampling the canonical statistics through $\psi_i$ assumes that the VAE latent is approximately equivariant to the scaling and translation of the layout.
The latent grid of the window has a stride of 8 pixels, so the latent position $p$ corresponds to the pixel center $8(p+1/2)$.
We map this center into the canonical frame by $\psi_i^{-1}$, convert it to canonical latent coordinates and interpolate the mean bilinearly.
The aligned mean is given by
\begin{equation}
  \tilde\mu_i(p)=\sum_k\alpha_{ik}(p)\,\mu_{C_i}(k),
  \label{eq:supp_warp_mean}
\end{equation}
where $\alpha_{ik}(p)$ denotes the bilinear weight of the neighboring canonical node $k$.
The aligned variance is that of this weighted sum of independent coordinates and is given by
\begin{equation}
  \tilde q_i(p)=\max\Big(\sum_k\alpha_{ik}(p)^2\,q_{C_i}(k),\;q_{\min}\Big).
  \label{eq:supp_warp}
\end{equation}
The diagonal model drops the correlations between neighboring positions.
Since $0\le\alpha_{ik}(p)\le1$, the aligned variance is at most the interpolated variance $\sum_k\alpha_{ik}(p)\,q_{C_i}(k)$.
The two are equal on the canonical grid, and the aligned variance is a quarter of the interpolated one at the center of a cell.
Therefore, off-grid positions receive narrower references and are drawn more strongly toward the mean.
When boundary clamping maps several neighbors to the same canonical node, we merge their weights before squaring.

Let $c_i(p)\in[0,1]$ denote the fraction of the latent cell $p$ covered by $F_i$, and let $\chi_i(p)\in\{0,1\}$ indicate the valid support of $\psi_i$.
The spatial weight is given by
\begin{equation}
  m_i(p)=\frac{c_i(p)\,\chi_i(p)}{\max\big(1,\sum_{i'\in b}c_{i'}(p)\big)},
  \label{eq:supp_weights}
\end{equation}
so $m_i(p)\ge0$ and $\sum_{i\in b}m_i(p)\le1$ hold at every position, including those where the full regions of several members overlap.
The weight is shared across the latent channels.
It uses $F_i$ instead of $V_i$, consistent with the condition $\kappa_b$, so the whole object is guided even where another block covers it.

% =====================================================================
\section{Experimental Details}
\label{sec:supp_exp}

This section provides the details of the datasets and budgets, the student training, the synthesis configuration, the adapted baselines and the evaluation protocols.

\subsection{Datasets, Budgets and VOC Masks}
\label{sec:supp_voc}

We distill COCO train2017 and evaluate on COCO val2017.
COCO provides an instance mask for every annotated object, and we discard the crowd annotations.
For PASCAL VOC, we distill the 2007 and 2012 trainval sets and evaluate on the 2007 test set.
\Cref{tab:supp_budgets} lists the canvas numbers $M$ and the instance budgets $B$ of all settings in the main paper, together with the sizes of the candidate pools and the degree of reference sharing.

Since VOC provides instance masks for only part of its images, we obtain the mask of every annotated object with SAM prompted by its ground-truth box.
In this way, SAM uses no information beyond the existing box annotations.
These masks enter Geometry Distillation in the same way as the COCO masks, and the boxes on the canvases are the tight boxes of the visible masks.
\Cref{fig:supp_masks}b shows these masks on VOC canvases.

\begin{table*}[t]
  \centering
  \small
  \caption{Budgets and reference sharing of all settings. $M$ and $B$ denote the number of canvases and the instance budget, respectively. The clusters are computed once per dataset on the full candidate pool. The last three columns report the mean number of selected instances per cluster and the fractions of the candidates and of all valid non-crowd instances that never appear on a canvas, respectively.}
  \label{tab:supp_budgets}
  \begin{tabular}{llrrrrrrr}
    \toprule
    Dataset & Budget & $M$ & $B$ & Candidates & Clusters & Inst./cluster & Unseen cand. & Unseen inst. \\
    \midrule
    COCO & $0.5\%$ & 592     & 25{,}038 & 426{,}445 & 787 & 31.81 & 94.13\% & 97.05\% \\
    COCO & $1\%$   & 1{,}184 & 50{,}000 & 426{,}445 & 787 & 63.53 & 88.28\% & 94.12\% \\
    \midrule
    VOC  & $0.5\%$ & 83      & 3{,}520  & 28{,}952  & 94  & 37.45 & 87.84\% & 92.55\% \\
    VOC  & $1\%$   & 166     & 7{,}026  & 28{,}952  & 94  & 74.74 & 75.73\% & 85.12\% \\
    \bottomrule
  \end{tabular}
\end{table*}

\subsection{Student Training}
\label{sec:supp_students}

We follow the evaluation protocol of OD$^3$ without modification.
For each detector, the students of all internally constructed datasets are trained with the same schedule and with PKD feature distillation from the same fixed teacher trained on the full training set.
The Full row of Table~1 of the main paper reports the Faster R-CNN teacher, which also provides the detector evidence for the candidate pool.

The baseline results in Table~1 of the main paper, including those of the coreset selection methods, are taken from their respective source papers and retain their original evaluation protocols.
The OD$^3$ results in Tables~2 and~3 and Figures~4 and~6 of the main paper require additional students or per-category and per-image predictions.
For these results, we train the students on the released OD$^3$ data under the same protocol.

\subsection{Synthesis Configuration}
\label{sec:supp_synthesis}

We render the contours at half of the window resolution and use the category names as the region texts.
The global prompt is the fixed sentence \emph{A photorealistic arrangement of real objects, preserving their shapes, poses and visible relationships, with natural colors and soft neutral lighting.}
The FLUX guidance scale is set to 3.5, and true classifier-free guidance is disabled.
The canonical latents use the posterior mode of the VAE with the standard FLUX latent shift and scale, and spherical $k$-means uses two initializations with at most 50 iterations each.
We sample each block with 32 Euler steps of the flow-matching scheduler with its resolution-dependent shift, starting from $\sigma_0=0.65$, and set $\lambda=0.25$ and $q_{\min}=10^{-4}$ for GFG.
The network runs in BF16, whereas the sampling state, the reference predictions and the fusion are computed in FP32.
GFG adds no network evaluation to Regen.
Its additional cost per step consists of the element-wise reference predictions and the fusion of Eq.~(8) of the main paper.
We derive the noise and condition seeds from the canvas identifier and the block index, so a change of the depth order does not change the random inputs.
All internal comparisons share these inputs.

\subsection{Adapted Baselines and Controls}
\label{sec:supp_baselines}

This section details the strategies and controls compared in Tables~3 to~5 of the main paper.
All generative strategies in Table~5 share the layouts, block windows, conditions, cluster assignments, Gaussian noise, number of steps, write-back and student training of GFG.
As a result, they differ only in the initialization and the guidance.
The D$^4$M-style and MGD$^3$-style baselines are mechanism-level controls implemented in our detection framework, and they do not reproduce the original classification methods.

\paragraph{Unguided Seg2Any.}
This baseline starts from pure noise with $\sigma_0=1$ and samples without reference guidance by setting $\lambda=0$, so it uses neither the anchor nor the references.

\paragraph{D$^4$M-style Initialization.}
This baseline replaces the anchor latent inside the support of every member by the spatially mapped cluster prototype and keeps the latent of the gray background elsewhere.
The prototype is the aligned cluster mean $\tilde\mu_i$ of \cref{eq:supp_warp_mean}.
The baseline then noises the result and samples without reference guidance.
% The noise level of this baseline is left unspecified here. If it is confirmed to be the same sigma_0 = 0.65 as GFG, this can be stated.

\paragraph{MGD$^3$-style Guidance.}
This baseline keeps the anchor and replaces the reference prediction $\bar x_{G,i}$ in Eq.~(7) of the main paper by the aligned cluster mean $\tilde\mu_i$.
By \cref{prop:supp_reference}(ii), this replacement is the zero-variance limit of Eq.~(4) of the main paper.
The baseline uses the same strength $\lambda=0.25$ and the same spatial weights $m_i$ as GFG.
The full variant guides all 32 steps and is therefore the zero-variance limit of GFG.
The stopped variant guides only the first 24 steps and then continues without guidance, as Regen does.
In the main paper, Table~5 reports the stopped variant, and Figure~5 shows both variants.

\paragraph{Independent-instance Control.}
The second and third rows of Table~3 of the main paper compare independent instances with real instance blocks under Copy.
The control uses the same source instances as the block geometry, so the category counts and the instance budget are also the same.
It places every instance as a single-member block on the same number of canvases with the same gray background.
The distributions of scale and visible area are matched to those of the block geometry.
Touching neighbors are therefore no longer kept together, and occlusion between instances arises only from the layout.

\paragraph{Sources of the Reference Statistics.}
Table~4 of the main paper keeps the cluster partition and the geometry, anchors and seeds of the $0.5\%$ COCO set.
It changes only the members of each cluster from which the mean and the variance of \cref{eq:supp_mean,eq:supp_stats} are computed.
The three variants use the candidates selected at this budget, the candidates not selected and all candidates, respectively, and the last variant is GFG.
Since every rendered instance is selected, every cluster that guides an instance has at least one selected member.
Under the first variant, the variance of a cluster with a single selected member equals the floor $q_{\min}$.
% Confirm that Eq. (supp_stats) is applied unchanged to each subset.

\subsection{Evaluation Protocols}
\label{sec:supp_protocols}

This section gives the protocols of the instance-level metrics defined in the main paper.

\paragraph{Diversity and Coverage.}
Div and CovDist are computed on the $L_2$-normalized 768-dimensional pooled features of the frozen SigLIP image encoder (\texttt{google/siglip-base-patch16-224}).
This encoder is independent of our method, as it is used neither for candidate selection nor for generation.
As defined in the main paper, both metrics are macro-averaged over the same clusters for all strategies.

\paragraph{Category Recognizability.}
Valid is the fraction of instances whose category the same frozen SigLIP model recognizes in a zero-shot manner.
For each instance, we set the pixels outside its visible mask to the gray value $(114,114,114)$ and crop the image to the tight box of the visible mask.
The crop is padded to a centered square and resized to $224\times224$ pixels with bicubic interpolation.
The text feature of each of the 80 COCO categories is computed from the prompt \texttt{a photo of a \{class\}.}
The predicted category is the one whose text feature has the highest cosine similarity with the normalized image feature.
A prediction is counted as correct if it equals the annotated category, and no confidence threshold is applied.
All strategies are evaluated on the same instances, and Copy gives the reference level of Valid on real pixels.

\paragraph{Mask Agreement.}
mIoU is the mean IoU between the visible mask $V_i$ and the mask predicted by SAM~3.1 (\texttt{sam3.1\_multiplex.pt}), averaged with equal weights over the same instances as Valid.
We prompt SAM~3.1 with the tight box of $V_i$ and one positive point, which is the pixel of $V_i$ with the largest Euclidean distance to the pixels outside $V_i$.
Each instance is predicted independently on the complete generated canvas.
The prediction is clipped neither to the prompt box nor to $V_i$, and an empty prediction is counted as an IoU of zero.

\paragraph{Qualitative Examples.}
Figure~5 of the main paper shows six instances of one cluster of the $0.5\%$ COCO set.
Every row shows the same instances rendered from the same geometry, conditions and noise.

% =====================================================================
\section{Additional Results}
\label{sec:supp_results}
\label{sec:supp_cross}

To further examine the cross-detector transfer, we report all six COCO metrics of the YOLO26 students in Table~2 of the main paper.
As shown in \Cref{tab:supp_cross}, GFG improves over OD$^3$ by $3.2$ AP, and the gain is larger at AP$_{75}$ than at AP$_{50}$, consistent with Table~1 of the main paper.
By object size, the gain comes from medium and large objects, where AP$_\mathrm{m}$ and AP$_\mathrm{l}$ improve by $3.2$ and $8.7$ points, respectively.
In contrast, AP$_\mathrm{s}$ decreases by $3.3$ points, probably because small objects are difficult to re-render at the token resolution of the generator.

\begin{table}[t]
  \centering
  \small
  \setlength{\tabcolsep}{4.5pt}
  \caption{Full COCO metrics of the YOLO26 student at the $0.5\%$ budget. The AP column repeats Table~2 of the main paper. The OD$^3$ student is trained on the released OD$^3$ data, and the GFG student is trained on the same distilled set as all other students. Best results are in bold.}
  \label{tab:supp_cross}
  \begin{tabular}{@{}lcccccc@{}}
    \toprule
    Method & AP & AP$_{50}$ & AP$_{75}$ & AP$_\mathrm{s}$ & AP$_\mathrm{m}$ & AP$_\mathrm{l}$ \\
    \midrule
    OD$^3$ & 13.9 & 22.8 & 14.8 & \textbf{8.5} & 21.5 & 13.7 \\
    GFG    & \textbf{17.1} & \textbf{25.8} & \textbf{18.6} & 5.2 & \textbf{24.7} & \textbf{22.4} \\
    \bottomrule
  \end{tabular}
\end{table}

% =====================================================================
\section{Limitations}
\label{sec:supp_limitations}

Our method has the following limitations.
First, as in generative dataset distillation for classification, rendering with a large generator is costly.
Detection raises this cost because a distilled set contains many more instances than images.
At the $1\%$ COCO budget, the $1{,}184$ canvases contain $50{,}000$ instances in $17{,}493$ blocks, or about 15 blocks per canvas.
Each block requires its own sampling run with 32 evaluations of the generator, although GFG itself adds only element-wise operations to Regen.

Second, small objects remain difficult to re-render.
The generator operates on tokens of $16\times16$ pixels, so a small instance covers only a few tokens.
For example, an instance at the minimum short side of 32 pixels spans only two tokens along this side.

Third, the reference model is a diagonal Gaussian.
It models neither the correlations between latent coordinates nor multiple modes within a cluster, so it describes the appearance variation of a cluster only approximately.

Finally, our method depends on several large pretrained models, namely FLUX.1-dev with the Seg2Any LoRA for rendering, DINOv3 for instance selection and SAM for the instance masks on VOC.
In future work, we will extend the reference model beyond diagonal Gaussians and improve the rendering of small objects.

\end{document}